%% file: fl_saliency.tex
\documentclass[10pt,twocolumn,letterpaper]{article}

\usepackage[table, svgnames, dvipsnames]{xcolor}
\usepackage{iccv}
\usepackage{times}
\usepackage{epsfig}
\usepackage{graphicx}
\usepackage{amsmath}
\usepackage{amssymb}
\usepackage{placeins}

\usepackage[T1]{fontenc}    
\usepackage{url}            
\usepackage{booktabs}       
\usepackage{amsfonts}       
\usepackage{nicefrac}       
\usepackage{microtype}      
\usepackage{bm}
\usepackage{makecell}
\usepackage{cellspace}

\usepackage{enumitem}
\usepackage{caption}
\usepackage{tabularx}

\usepackage{pifont}
\newcommand{\cmark}{\ding{51}}%
\newcommand{\xmark}{\ding{55}}%

\usepackage{array}
\usepackage{varwidth}

\usepackage{tikz}
\usetikzlibrary{positioning}

\tikzset{
	block/.style={
		draw,
		rectangle,
		align=center,
		rounded corners,
		minimum height=5ex,
		font=\sffamily,
	},
	conv/.style={
		block,
		fill=orange!25,
	},
	actv/.style={
		block,
		node distance=2ex,
		minimum height=3ex,
	},
	relu/.style={
	    actv,
		fill=red!25,
	},
	softmax/.style={
	    actv,
	    fill=green!25,
	},
	composite/.style={
		block,
		fill=blue!15,
	},
	pooling/.style={
		block,
		fill=magenta!25,
	},
	op/.style={
		block,
		fill=teal!25,
	},
	fork/.style={
	    circle,
	    fill,
	    inner sep=.25ex,
	},
}

\definecolor{tableau-c1}{RGB}{174,199,232}
\definecolor{tableau-c2}{RGB}{255,187,120}
\definecolor{tableau-c3}{RGB}{152,223,138}
\definecolor{tableau-c4}{RGB}{255,152,150}
\definecolor{tableau-c5}{RGB}{197,176,213}
\definecolor{tableau-c6}{RGB}{196,156,148}

\usepackage[pagebackref=true,breaklinks=true,letterpaper=true,colorlinks,bookmarks=false]{hyperref}

\iccvfinalcopy 

\def\iccvPaperID{****} 

\ificcvfinal\fi
\begin{document}

\title{Free-Lunch Saliency via Attention in Atari Agents}

\author{Dmitry Nikulin${}^{1}$\\
{\tt\small d.nikulin@samsung.com}
\and
Anastasia Ianina${}^{1}$\\
{\tt\small a.ianina@samsung.com}
\and
Vladimir Aliev${}^{1}$\\
{\tt\small vladimiralieva@gmail.com}
\and
Sergey Nikolenko${}^{1,2,3}$\\
{\tt\small sergey@logic.pdmi.ras.ru}\\
${}^1$Samsung AI Center, Moscow, Russia\\
${}^2$Steklov Institute of Mathematics at St. Petersburg, Russia\\
${}^3$Neuromation OU, Tallinn, Estonia\\
}

\maketitle
\thispagestyle{empty}

\begin{abstract}
We propose a new approach to visualize saliency maps for deep neural network models and apply it to deep reinforcement learning agents trained on \emph{Atari} environments. Our method adds an attention module that we call FLS (Free Lunch Saliency) to the feature extractor from an established baseline~\cite{mnih_nature_2015}. This addition results in a trainable model that can produce saliency maps, i.e., visualizations of the importance of different parts of the input for the agent's current decision making. We show experimentally that a network with an FLS module exhibits performance similar to the baseline (i.e., it is ``free'', with no performance cost) and can be used as a drop-in replacement for reinforcement learning agents. We also design another feature extractor that scores slightly lower but provides higher-fidelity visualizations. In addition to attained scores, we report saliency metrics evaluated on the Atari-HEAD dataset of human gameplay.
\end{abstract}

\section{Introduction}

Reinforcement learning (RL) models train agents that process input information from the environment in order to learn and implement a policy that maximizes the desired reward~\cite{SB18}. Over the last decade, reinforcement learning has shifted towards \emph{deep} reinforcement learning, where policies and/or models of the environment are modeled with deep neural networks, with impressive results achieved across a wide variety of tasks, from game playing to robotics~\cite{8103164,DBLP:journals/corr/abs-1811-12560}. Unfortunately, both RL and deep learning in general suffer from poor \emph{interpretability}: it is hard to explain why a deep neural network performs as it does, hard to understand why an RL agent has assigned a specific value to a given state, and harder yet when these two inherently ``black-box'' methods come together~\cite{8397411}. On the other hand, interpretability is a key feature for many critical applications, especially in robotics. This motivates the design of interpretable RL agents and modifying existing architectures for easier interpretation.

In this work, we concentrate on RL problems where the environment is represented by an image; a common example is given by \emph{Atari} game environments popularized for deep RL in~\cite{mnih_atari_2013}. In this context, interpretability often comes in the form of \emph{saliency maps}, i.e., values of how important individual pixels or small patches of the input image are for the Q-function or current policy decision. In the context of deep convolutional networks, saliency maps were introduced in~\cite{saliency_simonyan}, where gradients of the output category for image classification problems were visualized with respect to the input components (pixels). A map of such values provides some intuition of pixel importance: large absolute values of the gradients tell us that changes in those pixels can significantly affect the output category. Later works shifted the focus from trying to interpret already trained models towards building interpretability into the models themselves, usually with some form of an \emph{attention mechanism}~\cite{bahdanau_attention_2014}; we give a survey of these and other methods in Section~\ref{sec:related}. However, adding built-in interpretability often leads to a significant decrease in the actual rewards obtained by the resulting RL agents, i.e., usually one can have either state of the art performance or interpretability but not both.

We aim to explore the possibility of having both interpretable visualizations and high rewards at the same time. We introduce a visualization layer used to change the feature extractor's architecture in such a way that it learns visualizations as a side effect of training. We conduct a comprehensive experimental study on six \emph{Atari} environments: BeamRider, Breakout, MsPacman, SpaceInvaders, Enduro, and Seaquest. As a result of applying soft attention to \emph{Atari} gameplay screenshots (adding a new layer), the resulting agent is able to localize its attention and generate saliency maps in the process of training. These saliency maps can be used to understand how the agent learns. Moreover, they provide a close match to human attention. To evaluate this effect, we have used the Atari Human Eye-Tracking and Demonstration Dataset (Atari-HEAD)~\cite{DBLP:journals/corr/abs-1903-06754} as the ground truth, measuring the similarity between saliency maps and human eye movements across three metrics: normalized scanpath saliency (NSS), KL divergence, and shuffled AUC.

Thus, we present a useful and interpretable way for visualization. Unlike other ways of visualizing the agents' performance, such as Jacobians or reward curves~\cite{wang_dueling_2015}, saliency maps can allow even non-experts to draw reasonable conclusions. Moreover, it provides an easy way for debugging agents and interpreting the policy.

The paper is organized as follows. In Section~\ref{sec:related}, we provide an overview of different ways to improve the interpretability of deep RL models and agents. Section~\ref{sec:main} introduces our attention-based modifications to the baseline RL agents and previously developed models. A comprehensive evaluation study of our model against these competitors is provided in Section~\ref{sec:eval}. We speculate on potential avenues for future research and conclude the paper in Section~\ref{sec:concl}.

\section{Related work}\label{sec:related}

In this section, we review the recent interpretability and saliency studies in deep reinforcement learning (RL). Since we had been unable to find comprehensive surveys of existing work to reference during literature review, we decided to make this section into such a review.
The deep RL revolution was started by~\cite{mnih_atari_2013}, who successfully used deep CNNs trained using Q-learning to play \emph{Atari} games from raw pixels. In a follow-up work~\cite{mnih_nature_2015} they were also the first to attempt to interpret trained models by applying t-SNE to network hidden states. They used a single architecture to play all of the environments. We show its feature extractor in Figure~\hyperref[fig:architectures]{2.a}, with the corresponding \emph{Sparse block} in Figure~\hyperref[fig:conv]{1.a}.
We broadly divide work on interpreting deep RL models and deep neural networks in general into two major categories: \emph{post-hoc} and \emph{built-in} saliency. Next, we briefly review the former and provide a comprehensive survey of the latter.

\begin{description}[leftmargin=0pt]
    \item[Post-hoc saliency] includes approaches where additional techniques intended for interpretation follow an already trained model, without affecting the model and the training process itself.

          As we have already noted, saliency maps were first introduced in deep learning in the context of image classification by~\cite{saliency_simonyan}, who proposed to visualize gradients of output category with respect to input image. This general approach was extended in many later works, most notably with \emph{guided backpropagation}~\cite{springenberg2014striving}, \emph{integrated gradients}~\cite{sundararajan2017axiomatic}, \emph{Grad-CAM}~\cite{selvaraju2017grad}, \emph{LRP}~\cite{bach2015lrp}, and \emph{DeepLIFT}~\cite{shrikumar2017deeplift}. The current state of the art approaches are \emph{SmoothGrad}~\cite{smilkov2017smoothgrad} and \emph{VarGrad}~\cite{adebayo2018vargrad}, which were studied theoretically in \cite{seo2018noise} and found to be best in a recent comprehensive evaluation~\cite{hooker2018evaluating}.

          In the context of deep RL in particular, studies of interpretability were pioneered by~\cite{mnih_nature_2015} who applied t-SNE to a CNN network playing Atari games; later, Jacobians of value and advantage streams with respect to input images were used for the same purpose in~\cite{wang_dueling_2015}. The work~\cite{zahavy2016graying} investigated t-SNE embeddings in more depth. In~\cite{greydanus_vauaa_2017}, saliency maps for Atari agents trained with A3C were visualized by blurring different parts of input images and measuring the $L^2$ difference between actor and critic outputs; such an approach is computationally expensive but the resulting images are clearer than Jacobians. Recently, the work~\cite{weitkamp_rationalizations_2019} took a closer look at a slightly modified version of Grad-CAM in the context of deep RL on Atari games.

    \item[Built-in saliency] refers to approaches where interpretability follows from special constructions added to the models themselves. There is already a significant number of works in this direction. We give a brief summary of their main features in Table~\ref{table:papers} and describe them in more detail below.
\end{description}

\begin{table*}
    \centering 
    \setlength{\tabcolsep}{3pt}
    \footnotesize
    \begin{tabular}{c|lp{.9cm}p{.7cm}p{1.4cm}p{.9cm}p{2.7cm}p{.9cm}p{.8cm}p{1.5cm}p{3.4cm}}
    Ref                              & Algorithm & Custom env & \# envs (\emph{Atari}) & Based on & Train- able attn & Attention types                & Attention architecture        & Sum- pool & Vis. method       & Metrics                  \\ \hline
    \cite{sorokin_darqn_2015}         & DRQN      & \xmark     & 5  & \cite{hausknecht_drqn_2015} & \cmark & Soft and hard self-attention             & Fig.~\hyperref[fig:conv]{1.d} & \cmark   & Upscale            & Reward                   \\ \hline
    \cite{chen_observe_2017}          & DRQN      & \xmark     & 3  & \cite{hausknecht_drqn_2015} & \cmark & Soft temporal and spatial self-attention & Fig.~\hyperref[fig:conv]{1.d} & \cmark   & Upscale*           & Reward                   \\ \hline
    \cite{mousavi_where_to_look_2016} & DRQN      & \xmark     & 5  & \cite{hausknecht_drqn_2015} & \cmark & Soft self-attention                      & Fig.~\hyperref[fig:conv]{1.e} & \cmark   & Upscale*           & NSS, AUC                 \\ \hline
    \cite{choi_multi-focus_2017}      & DQN       & {\scriptsize GridWorld} & 0 & Custom          & \cmark & Soft key-value                           & See~\cite{choi_multi-focus_2017} & \cmark & Raw               & Reward                   \\ \hline
    \cite{zhang_agil_2018}            & Imitation & \xmark     & 8  & \cite{mnih_nature_2015}     & \cmark & Soft, on human data                      & See~\cite{zhang_agil_2018}    & N/A      & Raw                & Reward, NSS, AUC, KL, CC \\ \hline
    \cite{yuezhang_initial_2018}      & A2C       & Catch      & 4  & \cite{mnih_nature_2015}     & \xmark & N/A                                      & N/A                           & N/A      & Opt. flow          & Reward                   \\ \hline
    \cite{yang_ltiaa_2018}            & Rainbow   & \xmark     & 8  & \cite{mnih_nature_2015}     & \cmark & Soft                                     & Fig.~\hyperref[fig:conv]{1.c} & \xmark   & Jacobian           & Reward                   \\ \hline
    \cite{manchin_works_2019}         & PPO       & \xmark     & 10 & \cite{mnih_nature_2015}     & \cmark & Soft self-attention                      & *                             & \xmark   & Grad-CAM + upscale & Reward                   \\ \hline
    \cite{deepmind_2019}              & IMPALA    & \xmark     & 57 & \cite{espeholt2018impala}   & \cmark & Soft key-value                           & See~\cite{deepmind_2019}      & \cmark   & Raw                & Reward                   \\ \hline
    \rowcolor{lightgray}
    Ours                              & PPO       & \xmark     & 6  & \cite{mnih_nature_2015}, custom & \cmark & Soft self-attention                  & Fig.~\hyperref[fig:conv]{1.f} & both     & $\mathrm{conv}^T$  & Reward, NSS, KL, sAUC    \\ \hline
    \end{tabular}
    \captionof{table}{Papers on ad-hoc saliency. Asterisks show cases where the original paper is vague on exact details.
    }
    \label{table:papers}
\end{table*}

One of the first notable models with saliency in RL via attention is DARQN~\cite{sorokin_darqn_2015} (Figure~\hyperref[fig:conv]{1.d}), a modification of DRQN~\cite{hausknecht_drqn_2015} augmented with spatial attention dependent on the recurrent state. The authors experimented with both soft attention (regular spatial attention) and hard attention (sampling from the distribution generated by the attention network). The results are ambiguous: while achieving high scores on Seaquest, they scored lower than baseline scores on Breakout. We compare our approach against a non-recurrent variation of their method, which we call DAQN.

In~\cite{chen_observe_2017}, a similar approach is augmented by experiments with temporal attention in addition to spatial attention. Despite the fact that temporal attention appears to improve performance, it cannot be used to obtain saliency maps, so we do not consider it in our work. The work~\cite{chen_observe_2017} is vague on the exact procedure for obtaining visualizations, although they look like simple upscalings of attention activations.

The work~\cite{mousavi_where_to_look_2016} refers to assessors in order to get areas of the input image (frame) which people are most likely to look at during a game. The collected data is used as a proxy for gaze positions. Comparing to previous works, the model is trained with a smaller attention module, and the resulting saliency maps are evaluated by measuring how similar they are to data collected from humans. The authors compare their approach with two methods for determining saliency that are not based on DL: Itti-Koch saliency model~\cite{itti_koch_1998} and Graph-Based Visual Saliency (GBVS)~\cite{Harel:2006:GVS:2976456.2976525}. They show that both of them are no better than random, while attention-based saliency significantly outperforms them. They also do not specify their visualization method precisely.

The work~\cite{zhang_agil_2018} also uses human attention data but brings a different approach: they have collected a dataset of human gaze positions with an eye tracker, trained an encoder-decoder architecture to predict human gaze positions using that data, and finally used it to train an agent with imitation learning, achieving promising results.

The approach of~\cite{yang_ltiaa_2018} is very similar to ours: their main modification is the addition of a custom attention block (Figure~\hyperref[fig:architectures]{2.c}) that they call RS (Region-Sensitive) module. They train the resulting model with Rainbow~\cite{DBLP:conf/aaai/HesselMHSODHPAS18} and show that the it learns to focus its attention on semantically relevant areas. The work~\cite{manchin_works_2019} tested variations of self-attention but used Grad-CAM to visualize saliency maps.

Inspired by selective attention models, the authors of~\cite{yuezhang_initial_2018} use optical flow between two frames as the attention map. They experiment with A2C on Atari games, conducting several experiments on DQN with and without attention and reporting moderate performance improvements on four Atari games (Breakout, Seaquest, MsPacman, and Centipede). However, they provide no metrics or ways to estimate the result quantitatively. The authors remark that more experimental data is needed to show the benefits of visual attention for deep RL models, stating that ``experiments on more games should be conducted to provide a more comprehensive evaluation for the effect of introducing visual attention''.

In~\cite{choi_multi-focus_2017}, an attention model is applied to a toy problem in single-agent and multi-agent settings. The authors show better performance compared to the DQN and report advantages of a multi-agent architecture over a single-agent one. Furthermore, they report 20\% better sample efficiency.

In a recent paper~\cite{deepmind_2019}, soft top-down attention in a recurrent model is used to force the agent to focus on task-relevant information. The resulting agent exhibits performance comparable to state of the art on Atari games.

\section{Our approach}\label{sec:main}

\def\picscale{.63}

\input{fig_architectures}

Our work continues the general idea of applying attention improvements to RL agents in order to get interpretability while hopefully not sacrificing performance. Unlike all works discussed in Section~\ref{sec:related}, we suggest a quantitative way to evaluate attention maps produced by the models, using the information from an eye-tracker recently made available in the Atari Human Eye-Tracking and Demonstration Dataset (Atari-HEAD)~\cite{DBLP:journals/corr/abs-1903-06754}. This allows us to compare multiple architectures and model setups in order to find the best way of introducing attention to deep RL models.

Contrary to some of the prior art, the \emph{Sparse FLS} architecture we propose in our work is a small, incremental change from the non-recurrent baseline. We conduct experiments using 5 random seeds across 6 Atari environments, and in all of our experiments, our model performs similarly to the baseline, while also producing visualizations. We also compute saliency metrics using the Atari-HEAD dataset~\cite{zhang_agil_2018} of human eye fixations captured using an eye tracker from human players playing Atari games, showing that neural models perform significantly better than random in approximating human gazes. Finally, we also propose a \emph{Dense FLS} architecture, which, despite attaining lower scores, generates significantly sharper images.

Deep RL models are notorious for their sensitivity to hyperparameters~\cite{henderson_2018_deep_rl_that_matters}. For this reason, we primarily investigate incremental changes to approaches that are known to work well~\cite{mnih_atari_2013,mnih_nature_2015} rather than attempt to build a new one from scratch. Since post-hoc saliency methods have been studied quite extensively (see Section~\ref{sec:related}), we develop a model that has built-in saliency. We turn to the idea of attention \cite{bahdanau_attention_2014}, and, borrowing elements from~\cite{noh_delf_2016}, integrate visual soft self-attention into a baseline model from~\cite{mnih_nature_2015}. We show details of the architectures in Figure~\ref{fig:architectures}.

Similar to~\cite{yang_ltiaa_2018} (Fig.~\hyperref[fig:architectures]{2.c} with the RS module detailed in Fig.~\hyperref[fig:conv]{1.c}; see Section~\ref{sec:related}), we add an extra self-attention module between convolutional and fully-connected layers. One of the main differences is that we use SoftPlus~\cite{softplus_dugas2001incorporating} as the activation function for the final layer of the FLS module. The choice of SoftPlus is inspired by~\cite{noh_delf_2016} and motivated by the fact that, in contrast to previous approaches~\cite{sorokin_darqn_2015,chen_observe_2017,mousavi_where_to_look_2016,yang_ltiaa_2018}, it does not use any normalization. We have verified experimentally that normalizing the output of SoftPlus makes the model perform worse (see Section~\ref{exp_performance}). We have also seen that adding sum-pooling makes the model perform worse.

In addition to the \emph{Sparse} block which is a part of the baseline model from~\cite{mnih_nature_2015}, we also propose a different convolutional block which we call \emph{Dense}, as shown in Fig.~\ref{fig:conv}. It is designed in such a way that the receptive fields and strides of neurons in its final layer are small, making visualizations crisper; but, as we will see in experiments, this comes at the cost of the achieved reward. This model can only reasonably fit in GPU memory if sum-pooling is applied after attention.

We visualize saliency maps generated by FLS modules by drawing receptive fields of all neurons from the final convolutional layer with intensity proportional to the activations of the corresponding neurons in the attention layer, which we implement via transposed convolution of the output of the FLS module with a unit kernel (i.e., a tensor filled with ones) with suitable kernel size, strides, and padding. This approach strikes the balance between bilinear upscaling of the attention activations and more mathematically sound but extremely noisy Jacobian of the input image with respect to some function of attention activations.

In the next section, we proceed to show direct experimental comparisons against the architecture from~\cite{yang_ltiaa_2018} trained with PPO (which we call RS-PPO) and a non-recurrent version of~\cite{sorokin_darqn_2015} (denoted DAQN).

\section{Experimental evaluation}\label{sec:eval}

\subsection{Setup}

We run all experiments on 6 \emph{Atari} games \cite{bellemare_atari_2013}: BeamRider, Breakout, MsPacman, SpaceInvaders, Enduro, and Seaquest. This is identical to the set of games reported in an early version of~\cite{yang_ltiaa_2018} (a later version added Frostbite) with the exception that we replaced Pong with Breakout because in Pong, the score is capped at 21, and it is relatively easy to train an agent that achieves perfect or near-perfect score, which trivializes many comparisons. We use environments provided by the \emph{OpenAI Gym} library~\cite{gym}, specifically their \texttt{NoFrameskip-v4} versions.

All agents were trained using the Proximal Policy Optimization (PPO) algorithm~\cite{schulman_ppo_2017} as implemented in the \emph{OpenAI Baselines} library~\cite{baselines}, with default hyperparameters. We ran 8 environments in parallel and capped the total number of environment steps at $5 \cdot 10^{7}$ to limit resource usage.

\begin{figure}[t]
    \centering
    \includegraphics[width=\linewidth]{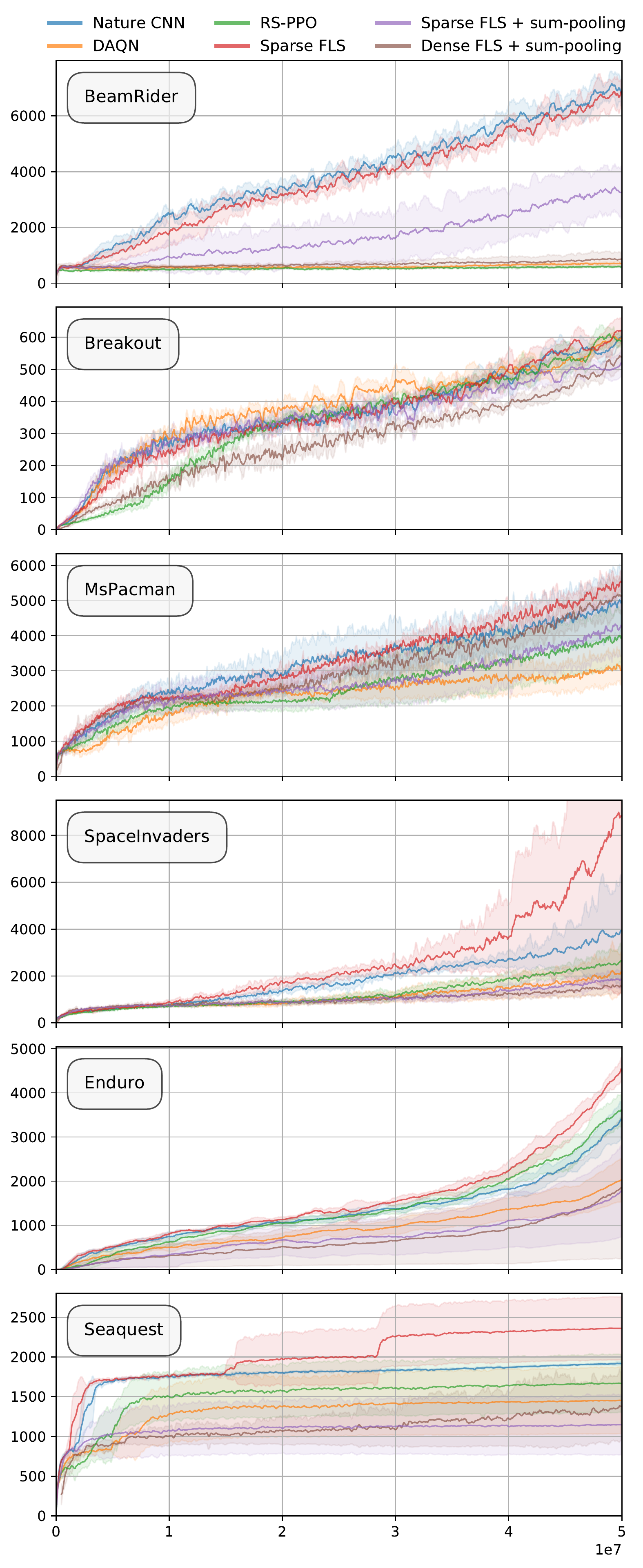}

    \caption{Reward curves during training. The horizontal axis shows the number of environment frames, the vertical axis shows the current reward.}
    \label{fig:averaged_curves}\vspace{-.4cm}
\end{figure}

For each environment and each architecture, we trained 5 agents with different random seeds. For every experiment, we recorded a smoothed curve of episode rewards the agents obtained during training. Reward curves in Figure~\ref{fig:averaged_curves} show the mean score and its standard deviation on every timestep. Specifically, we show the \texttt{eprewmean} metric reported by the Baselines library and computed as follows: during training, collecting experience is interleaved with agent updates. Experience is collected in 8 threads, which, under default hyperparameters, are executed for 128 steps at a time, totaling 1024 environment steps between agent updates. The \texttt{eprewmean} metric is the average reward for the last 100 episodes completed by the time the experience collection step is over. This includes episodes that are completed in steps prior to the current one. We have made the source code for the implementation of our models and reproducing the experiments available at~\url{https://github.com/dniku/free-lunch-saliency}.

\begin{table}[!t]\centering\footnotesize
    \begin{tabular}{lrr}
        \toprule
        Model                               & \# params     & \% of~\cite{mnih_nature_2015} \\
        \midrule
        \rowcolor{tableau-c1}
        Nature CNN \cite{mnih_nature_2015}  & 1{,}686{,}693 & 100\%                         \\
        \rowcolor{tableau-c2}
        DAQN \cite{sorokin_darqn_2015}      & 130{,}726     & 7.8\%                         \\
        \rowcolor{tableau-c3}
        RS-PPO \cite{yang_ltiaa_2018}       & 1{,}720{,}999 & 102.0\%                       \\
        \rowcolor{tableau-c4}
        Sparse FLS                          & 1{,}836{,}710 & 108.9\%                       \\
        \rowcolor{tableau-c5}
        Sparse FLS + sum-pooling            & 263{,}846     & 15.6\%                        \\
        \rowcolor{tableau-c6}
        Dense FLS + sum-pooling             & 280{,}358     & 16.6\%                        \\
        Sparse + FLS after first conv layer & 1{,}762{,}982 & 104.5\%                       \\
        Sparse + FLS after each conv layer  & 2{,}063{,}016 & 122.3\%                       \\
        \bottomrule
    \end{tabular}\vspace{-.1cm}

    \caption{Model sizes. Colors correspond to Fig.~\ref{fig:averaged_curves}.}\vspace{-.3cm}
    \label{table:model_sizes}
\end{table}

\subsection{Performance}
\label{exp_performance}

Figure~\ref{fig:averaged_curves} shows reward curves obtained during training. It can be seen that our \emph{Sparse FLS} architecture, which is the baseline model with an extra attention module, performs similarly to the baseline, while our \emph{Dense FLS} architecture achieves lower rewards. We evaluate each trained model $2^{13}=8192$ times on environments initialized with a previously unseen random seed.

\begin{table*}[!t] \footnotesize\centering
    \begin{tabular}{p{4.7cm}p{1.6cm}p{1.6cm}p{1.6cm}p{1.6cm}p{1.6cm}p{1.6cm}}
        \toprule
        Game                                      & BeamRider     & Breakout    & Enduro        & MsPacman      & Seaquest        & SpaceInvaders   \\
        \midrule
        \rowcolor{tableau-c1}
        Nature CNN \cite{mnih_nature_2015}        & 6949$\pm$2569 & 618$\pm$209 & 3808$\pm$1670 & 4874$\pm$1701 & 1920$\pm$37     & 3867$\pm$3627   \\
        \rowcolor{tableau-c2}
        DAQN \cite{sorokin_darqn_2015}            & 701$\pm$205   & 601$\pm$201 & 2182$\pm$1075 & 3111$\pm$1165 & 1453$\pm$420    & 2096$\pm$1554   \\
        \rowcolor{tableau-c3}
        RS-PPO \cite{yang_ltiaa_2018}             & 583$\pm$185   & 605$\pm$202 & 3851$\pm$1677 & 3943$\pm$1435 & 1670$\pm$372    & 2562$\pm$1339   \\
        RS-PPO \cite{yang_ltiaa_2018} w/o padding & 823$\pm$432   & 591$\pm$199 & 3658$\pm$1670 & 3950$\pm$1371 & 1710$\pm$379    & 2248$\pm$782    \\
        \midrule
        \rowcolor{tableau-c4}
        Sparse FLS                                & 6634$\pm$2361 & 624$\pm$211 & 5094$\pm$1876 & 5421$\pm$1517 & 2440$\pm$382    & 9359$\pm$13230  \\
        \rowcolor{tableau-c5}
        Sparse FLS + sum-pooling                  & 3356$\pm$1878 & 520$\pm$183 & 1917$\pm$1486 & 4317$\pm$1485 & 1150$\pm$385    & 1847$\pm$773    \\
        Sparse FLS + norm                         & 6584$\pm$2159 & 598$\pm$200 & 4524$\pm$1807 & 3409$\pm$1275 & 1161$\pm$348    & 11206$\pm$10441 \\
        Sparse FLS w/ $1 \times 1$ convs          & 6870$\pm$2413 & 621$\pm$207 & 4701$\pm$1880 & 4887$\pm$1589 & 2252$\pm$336    & 5673$\pm$6344   \\
        Sparse FLS w/ $\operatorname{SoftPlus}_2$ & 6697$\pm$2261 & 612$\pm$208 & 4854$\pm$1823 & 5242$\pm$1527 & 1908$\pm$29     & 6443$\pm$7684   \\
        Sparse FLS w/o final ReLU                 & 6777$\pm$2242 & 589$\pm$207 & 4814$\pm$1904 & 5049$\pm$1145 & 2013$\pm$185    & 2929$\pm$556    \\
        Sparse FLS w/o final ReLU + sum-pooling   & 1057$\pm$1061 & 480$\pm$158 & 2093$\pm$1469 & 3365$\pm$1292 & 2356$\pm$1874   & 1999$\pm$899    \\
        Sparse + FLS after first conv layer       & 7468$\pm$2645 & 640$\pm$212 & 4942$\pm$2053 & 4720$\pm$1455 & 2181$\pm$376    & 9395$\pm$13615  \\
        Sparse + FLS after each conv layer        & 6588$\pm$2348 & 633$\pm$217 & 5950$\pm$3739 & 4978$\pm$1461 & 2083$\pm$314    & 2855$\pm$194    \\
        \rowcolor{tableau-c6}
        Dense FLS + sum-pooling                   & 866$\pm$415   & 532$\pm$173 & 2114$\pm$2164 & 4977$\pm$1253 & 1368$\pm$517    & 1549$\pm$831    \\
        Dense FLS w/o final ReLU + sum-pooling    & 730$\pm$245   & 503$\pm$162 & 1292$\pm$1829 & 4879$\pm$1282 & 10052$\pm$11853 & 1316$\pm$813    \\
        \bottomrule
    \end{tabular}\vspace{-.1cm}

    \caption{Evaluation scores. We trained 5 models with different random seeds for each (game, architecture) pair. Each model was evaluated $8192$ times on environments initialized with a previously unseen random seed, with results aggregated across models. Colors correspond to Fig.~\ref{fig:averaged_curves}.}
    \label{table:eval_scores}
\end{table*}

The final results are shown in Table~\ref{table:eval_scores}.
In addition to testing the architectures discussed in Section~\ref{sec:main}, Table~\ref{table:eval_scores} also shows an ablation study and the results of testing several variations intended to verify our hypotheses.

First, we hypothesize that inferior performance of our \emph{Dense FLS} model is at least partially due to spatial sum-pooling; we test this hypothesis by training an intermediate model, one with a Sparse block and spatial sum-pooling. Experimental results validate our hypothesis, showing much lower results for this intermediate model compared with the regular Sparse-based model. The loss of performance from sum-pooling may be caused by the loss of spatial information (i.e., the position of the ball and paddle in Breakout). Indeed, the work~\cite{deepmind_2019} suggests to concatenate fixed Fourier basis vectors to the tensor before applying spatial sum-pooling. We have not investigated the effects of similar workarounds. Note, however, that our \emph{Sparse FLS} architecture has approximately 10\% more parameters than the baseline, while the addition of sum-pooling reduces the number of parameters in the model by nearly a factor of 7 (see Table~\ref{table:model_sizes}). Thus, in some applications sum-pooling may be a sensible choice in terms of the memory-performance trade-off.

Second, we hypothesize that normalizing the output of the FLS module makes the model perform worse, and train a model where we divide the output of the FLS module by its sum. This, again, reduces the performance, yielding evidence in favor of this hypothesis. Third, we tested the model with $3 \times 3$ convolutions replaced with $1 \times 1$. Although this change was beneficial in our early experiments, a full ablation study revealed that it actually makes little difference.

Fourth, note that the FLS module can learn to output a constant value of $\ln 2$ if all of its weights are set to zero. We hypothesize that since we multiply the output of convolutional blocks by the output of the FLS module, model performance can be improved if this constant is instead $1$. To test that, we replace SoftPlus with its base-2 equivalent:
$\mathrm{SoftPlus}_2(x)
    = \log_2 \left( 1 + 2^x \right)
    = \frac1{\log 2} {\mathrm{SoftPlus}(x \cdot \log 2)}$.
Table~\ref{table:eval_scores} shows that this change does not significantly affect model performance. We also showed that the non-linearity before the FLS module is essential: removing it and feeding the output of the final convolutional layer directly into the FLS module severely degrades performance.

Finally, we experimented with other positions for the FLS module. Experiments suggest that inserting it after the first convolutional layer does not significantly affect performance; however, the resulting images tend to be less coherent that the ones produced by our \emph{Sparse FLS} model. Similarly, inserting an instance of the module after each convolutional layer does not impact performance, but the images obtained by summing attention masks from each module tend to be very blurry (see Section~\ref{sec:eval:visualizations} for visualizations).

\begin{table}[t]
    \footnotesize
    \setlength{\tabcolsep}{1.5pt}
    \begin{tabular}{lccccc}
        \toprule
        \bf Model  & \bf Breakout           & \bf Enduro             & \bf MsPacman            & \bf Seaquest           & \bf SpaceInv.           \\
        \midrule
        \multicolumn{6}{c}{\textbf{Normalized Scanpath Saliency (NSS)}}\\\midrule
        \rowcolor{tableau-c2}
        DAQN       & 1.344{\tiny$\pm$0.114} & 0.586{\tiny$\pm$0.356} & 0.881{\tiny$\pm$0.100}  & 0.334{\tiny$\pm$0.096} & 1.899{\tiny$\pm$0.052}  \\
        \rowcolor{tableau-c3}
        RS-PPO     & 0.947{\tiny$\pm$0.107} & 0.922{\tiny$\pm$0.123} & 0.943{\tiny$\pm$0.081}  & 0.955{\tiny$\pm$0.111} & 1.775{\tiny$\pm$0.039}  \\
        \rowcolor{tableau-c4}
        Sparse FLS & 0.510{\tiny$\pm$0.055} & 1.588{\tiny$\pm$0.072} & 0.631{\tiny$\pm$0.029}  & 0.556{\tiny$\pm$0.109} & 1.664{\tiny$\pm$0.015}  \\
        \rowcolor{tableau-c5}
        Sparse+SP  & 0.747{\tiny$\pm$0.267} & 0.251{\tiny$\pm$0.369} & 0.621{\tiny$\pm$0.059}  & 0.286{\tiny$\pm$0.067} & 1.569{\tiny$\pm$0.043}  \\
        \rowcolor{tableau-c6}
        Dense+SP   & 1.385{\tiny$\pm$0.545} & 0.629{\tiny$\pm$0.689} & -0.136{\tiny$\pm$0.188} & 0.797{\tiny$\pm$0.265} & -0.230{\tiny$\pm$0.557} \\
        \midrule
        \multicolumn{6}{c}{\textbf{KL divergence}}\\\midrule
        \rowcolor{tableau-c2}
        DAQN       & 2.885{\tiny$\pm$0.072} & 4.025{\tiny$\pm$0.370} & 3.364{\tiny$\pm$0.054}  & 4.418{\tiny$\pm$0.155} & 2.880{\tiny$\pm$0.063}  \\
        \rowcolor{tableau-c3}
        RS-PPO     & 3.209{\tiny$\pm$0.101} & 3.368{\tiny$\pm$0.093} & 3.325{\tiny$\pm$0.053}  & 3.294{\tiny$\pm$0.094} & 2.979{\tiny$\pm$0.029}  \\
        \rowcolor{tableau-c4}
        Sparse FLS & 3.472{\tiny$\pm$0.031} & 3.180{\tiny$\pm$0.033} & 3.491{\tiny$\pm$0.018}  & 3.492{\tiny$\pm$0.072} & 3.145{\tiny$\pm$0.019}  \\
        \rowcolor{tableau-c5}
        Sparse+SP  & 3.481{\tiny$\pm$0.342} & 4.752{\tiny$\pm$0.614} & 3.535{\tiny$\pm$0.049}  & 4.281{\tiny$\pm$0.417} & 3.086{\tiny$\pm$0.017}  \\
        \rowcolor{tableau-c6}
        Dense+SP   & 3.070{\tiny$\pm$0.261} & 3.453{\tiny$\pm$0.325} & 4.075{\tiny$\pm$0.232}  & 3.553{\tiny$\pm$0.299} & 4.186{\tiny$\pm$0.445}  \\
        \midrule
        \multicolumn{6}{c}{\textbf{Shuffled Area-Under-Curve (Shuffled AUC)}}\\\midrule
        \rowcolor{tableau-c2}
        DAQN       & 0.758{\tiny$\pm$0.024} & 0.527{\tiny$\pm$0.013} & 0.604{\tiny$\pm$0.017}  & 0.502{\tiny$\pm$0.011} & 0.679{\tiny$\pm$0.015}  \\
        \rowcolor{tableau-c3}
        RS-PPO     & 0.654{\tiny$\pm$0.029} & 0.552{\tiny$\pm$0.018} & 0.629{\tiny$\pm$0.024}  & 0.629{\tiny$\pm$0.020} & 0.678{\tiny$\pm$0.011}  \\
        \rowcolor{tableau-c4}
        Sparse FLS & 0.530{\tiny$\pm$0.012} & 0.530{\tiny$\pm$0.015} & 0.520{\tiny$\pm$0.005}  & 0.453{\tiny$\pm$0.027} & 0.656{\tiny$\pm$0.010}  \\
        \rowcolor{tableau-c5}
        Sparse+SP  & 0.612{\tiny$\pm$0.071} & 0.536{\tiny$\pm$0.017} & 0.524{\tiny$\pm$0.013}  & 0.486{\tiny$\pm$0.010} & 0.660{\tiny$\pm$0.020}  \\
        \rowcolor{tableau-c6}
        Dense+SP   & 0.698{\tiny$\pm$0.105} & 0.579{\tiny$\pm$0.107} & 0.610{\tiny$\pm$0.121}  & 0.693{\tiny$\pm$0.097} & 0.419{\tiny$\pm$0.110}  \\
        \bottomrule
    \end{tabular}
    \captionof{table}{Saliency metrics. Comparing against DAQN~\cite{sorokin_darqn_2015} and RS-PPO~\cite{yang_ltiaa_2018}. Sparse/Dense+SP denotes Sparse/Dense FLS with sum-pooling. Colors correspond to Fig.~\ref{fig:averaged_curves}.}
    \label{table:saliency_metrics}\vspace{-.3cm}
\end{table}

\subsection{Saliency metrics}

Saliency metrics estimate how well the saliency maps generated by a model approximate human eye fixations on the same images. We used the Atari-HEAD dataset of human actions and eye movements recorded while playing Atari videos games~\cite{zhang_agil_2018} in order to compare saliency maps generated by our models with human eye fixations. The dataset consists of $44$ hours of gameplay data from 16 games and a total of $2.97$ million demonstrated actions.

Computing the metrics is rather nontrivial due to a complex image preprocessing stack in the Baselines library whose design follows that of~\cite{mnih_atari_2013,mnih_nature_2015} and is widely regarded as standard. For the particular task of computing saliency metrics, the most relevant are the following steps. First, images are downscaled from $160 \times 210$ to $84 \times 84$ and converted to grayscale. Second, only two out of every four subsequent images are retained, and they are combined into one image with pixel-wise maximum. In the resulting stream, images are batched together in groups of size 4. Therefore, disregarding batch size, the neural network takes as input tensors of shape $84 \times 84 \times 4$, each of which corresponds to 8 images in the original stream. For saliency metrics, we took the union of all eye fixations in the corresponding 8 images as ground truth fixations for every frame. The saliency map was generated by applying transposed convolution to the FLS module activations as shown in Section~\ref{sec:main} and then upscaling the resulting map from $84 \times 84$ to $160 \times 210$. We averaged per-frame metrics over gameplay recordings, dropping frames with undefined metrics.

Following~\cite{riche2013saliency}, we computed three metrics: normalized scanpath saliency (NSS)~\cite{metrics_peters2005components}, KL divergence, and shuffled AUC~\cite{metrics_borji2012quantitative}. Throughout this section, we assume that the saliency map and fixation map are matrices of the same size; the saliency map contains floating-point values (output of the FLS module), while the fixation map contains integers, i.e., how many times an eye fixation was registered in a pixel while the human was viewing the image.

NSS measures the extent to which pixels with eye fixations are more prominent in saliency maps compared to other pixels. First, the saliency map is normalized to have zero mean and unit variance (NSS is undefined for saliency maps with zero variance). Then, the values of the pixels are averaged with weights equal to the number of fixations for the corresponding pixel:
\begin{equation}
    \operatorname{NSS}({f}, {s}) = \frac {\sum_{{i}, {j}} {f[i, j]} \cdot \widehat{{s}} {[i, j]}} {\sum_{{i}, {j}} {f[i, j]}},
\end{equation}
where ${s}$ is the saliency map, ${f}$ is the fixation map, and $\widehat{{s}}$ is the normalized saliency map.

Kullback-Leibler (KL) divergence is a pseudometric between probability distributions. It does not work well for discrete distributions such as a fixation map, which may have, e.g., non-intersecting supports~\cite{zhang_agil_2018}; therefore, before computing the KL divergence we blur fixation maps with Gaussian blur and $\sigma=5$, a value also used in~\cite{greydanus_vauaa_2017}:
\begin{equation}
    \operatorname{D_{KL}}({f}\|{s}) = \sum_{{i}, {j}} \overline{{f}} {[i, j]} \cdot \log \dfrac{\overline{{f}} {[i, j]}}{\overline{{s}} {[i, j]}},
\end{equation}
where $\overline{{s}}$ is the saliency map normalized to $\left[ 0 .. 1 \right]$, and $\overline{{f}}$ is the fixation map after blurring and a similar normalization.

Shuffled AUC (Area Under Curve) is a metric specifically designed for measuring the quality of saliency maps. It takes into account that the distribution of real fixations is skewed. The metric operates on a per-pixel level, regarding a saliency map as a prediction of the probability that there is an eye fixation in each pixel. As the name suggests, it computes the AUC by taking true fixations as true positives. However, as true negatives it takes real fixations for other frames in the same dataset. Shuffled AUC compensates for dataset bias by scoring a center prior at chance which implies that a model with more central predictions will have lower sAUC score than a model with predictions closer to the edges. Similar to AUC, shuffled AUC equal to $1$ indicates that the saliency model is perfect while being equal to $0.5$ means random predictions from the ground truth.

We summarize our experimental results in Table~\ref{table:saliency_metrics} (there is no BeamRider because it is not included in Atari-HEAD). Somewhat surprisingly, these experiments show that while all models perform better than random (an observation also made in~\cite{mousavi_where_to_look_2016}), no model can be singled out as a clear winner. \emph{Dense FLS} has the highest variance, which may be explained by the fact that the space of saliency distributions it is able to generate is larger than that of any other model.

\subsection{Visualizations}\label{sec:eval:visualizations}

\begin{figure*}[!t]
    \centering\footnotesize
    \setlength{\tabcolsep}{0pt}
    \begin{tabular}{cccccccc}
        \includegraphics[width=0.12\linewidth]{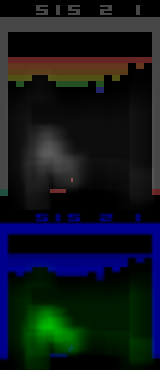}
        \label{fig:gameplay:breakout_dense_1}
        &
        \includegraphics[width=0.12\linewidth]{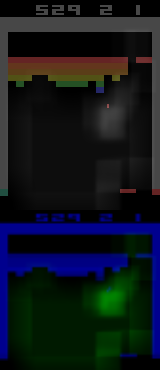}
        \label{fig:gameplay:breakout_dense_2}
        &
        \includegraphics[width=0.12\linewidth]{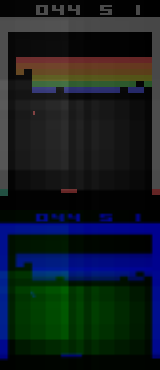}
        \label{fig:gameplay:breakout_sparse_1}
        &
        \includegraphics[width=0.12\linewidth]{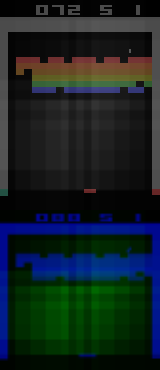}
        \label{fig:gameplay:breakout_sparse_2}
        &
        \includegraphics[width=0.12\linewidth]{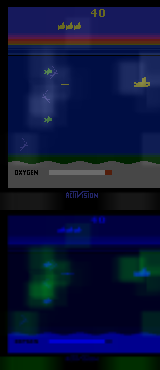}
        \label{fig:gameplay:seaquest_dense_1}
        &
        \includegraphics[width=0.12\linewidth]{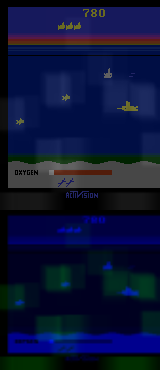}
        \label{fig:gameplay:seaquest_dense_2}
        &
        \includegraphics[width=0.12\linewidth]{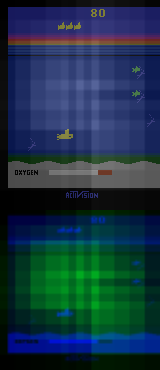}
        \label{fig:gameplay:seaquest_sparse_1}
        &
        \includegraphics[width=0.12\linewidth]{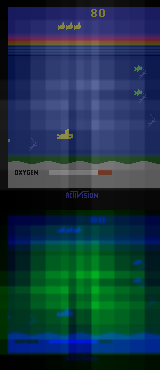}
        \label{fig:gameplay:seaquest_sparse_2}
        \\
        (a) & (b) & (c) & (d) & (e) & (f) & (g) & (h) \\
    \end{tabular}
    \caption{Game visualizations: (a-d) Breakout; (e-h) Seaquest; (a,b,e,f) \emph{Dense FLS}; (c,d,g,h) \emph{Sparse FLS}.}
    \label{fig:gameplay}\vspace{.3cm}
\end{figure*}

Figure~\ref{fig:gameplay} illustrates the information that can be gained via saliency maps. The top part of every image contains raw observations, while the blue channel of the bottom part shows preprocessed images as they are fed into the neural network. Saliency maps produced by the models are drawn in white for raw observations and in green for the preprocessed ones.

In general, Figure~\ref{fig:gameplay} shows that \emph{Dense FLS} produces crisp visualizations that are easy to interpret, but its performance is inferior to \emph{Sparse FLS}, which yields very coarse maps.

Figs.~\ref{fig:gameplay}(a)-(b) show \emph{Dense FLS} digging a tunnel through blocks in Breakout. The model focuses its attention on the end of the tunnel as soon as it is complete, suggesting that it sees shooting the ball through the tunnel as a good strategy.

Figs.~\ref{fig:gameplay}(c)-(d) depict the same concept of tunneling performed by the \emph{Sparse FLS} model. Note how it focuses attention on the upper part of the screen after destroying multiple bricks from the top. This attention does not go away after the ball moves elsewhere (not shown in the images). We speculate that this is how the agent models tunneling: rather than having a high-level concept of digging a tunnel, it simply strikes wherever it has managed to strike already.

Figs.~\ref{fig:gameplay}(e)-(f) illustrate how the \emph{Dense FLS} model playing Seaquest has learned to attend to in-game objects and, importantly, the oxygen bar at the bottom of the screen. As the oxygen bar is nearing depletion, attention focuses around it, and the submarine reacts by rising to refill its air supply.

Figs.~\ref{fig:gameplay}(g)-(h) are two consecutive frames where an agent detects a target appearing from the left side of the screen. The bottom part of the screenshots shows how attention in the bottom left corner lights up as soon as a tiny part of the target, only a few pixels wide, appears from the left edge of the screen. In the next frame, the agent will turn left and shoot the target (not shown here). However, the agent completely ignores targets in the top part of the screen, and its attention does not move as they move (also not shown).

\begin{figure*}[!t]
    \centering\footnotesize
    \setlength{\tabcolsep}{0pt}
    \begin{tabular}{ccc}
        \includegraphics[width=0.33\linewidth]{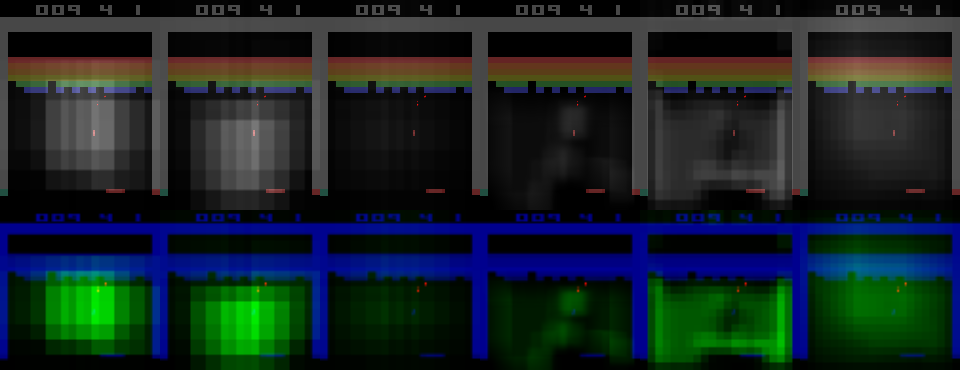}
        \label{fig:fixations:breakout_1}
        &
        \includegraphics[width=0.33\linewidth]{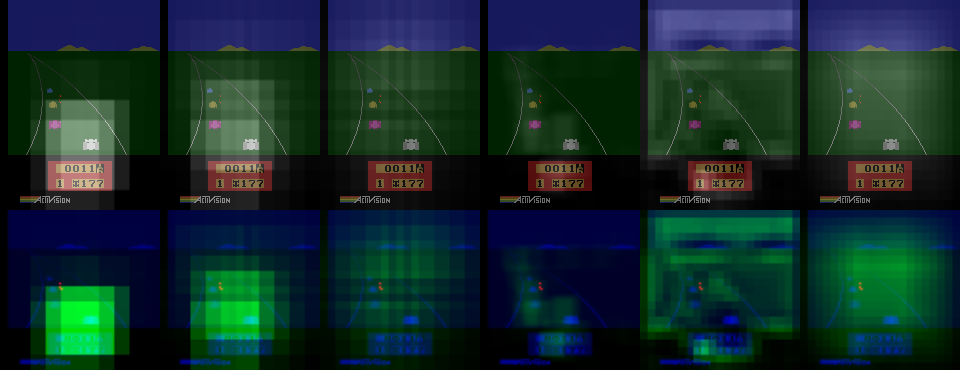}
        \label{fig:fixations:mspacman_1}
        &
        \includegraphics[width=0.33\linewidth]{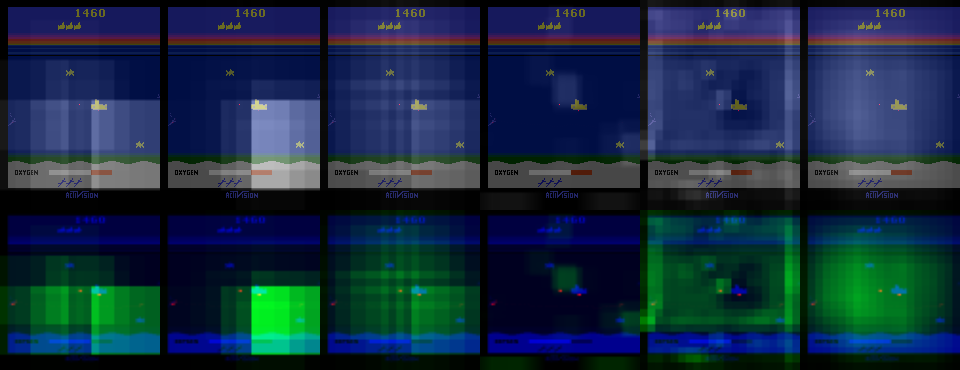}
        \label{fig:fixations:seaquest_1}
        \\ (a) & (b) & (c) \\
        \includegraphics[width=0.33\linewidth]{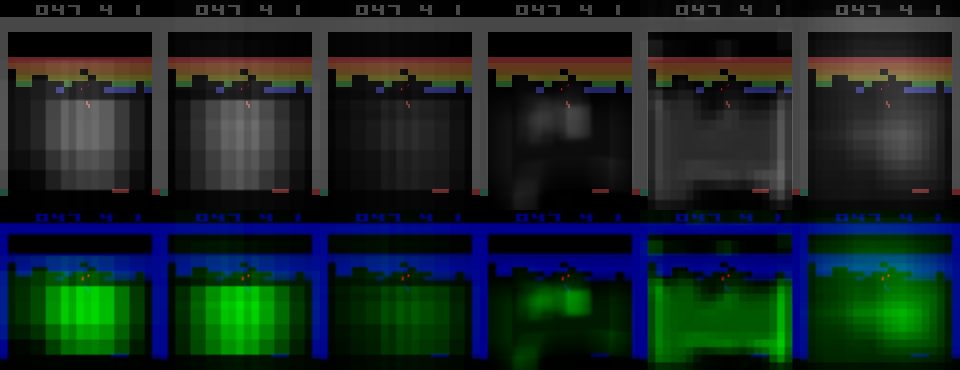}
        \label{fig:fixations:breakout_2}
        &
        \includegraphics[width=0.33\linewidth]{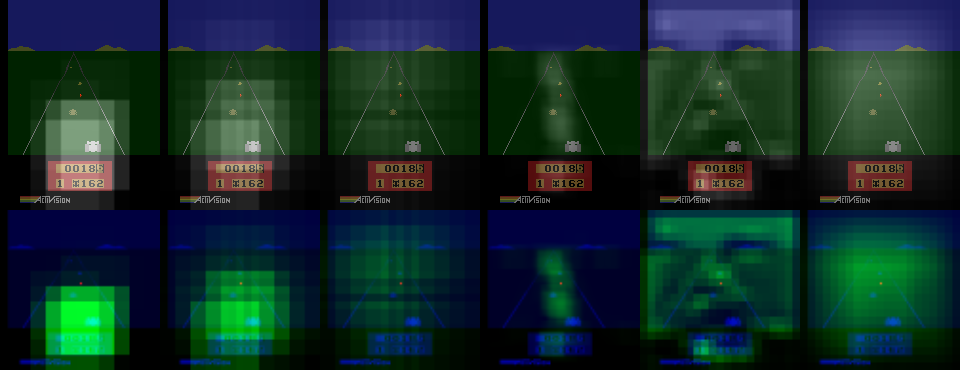}
        \label{fig:fixations:mspacman_2}
        &
        \includegraphics[width=0.33\linewidth]{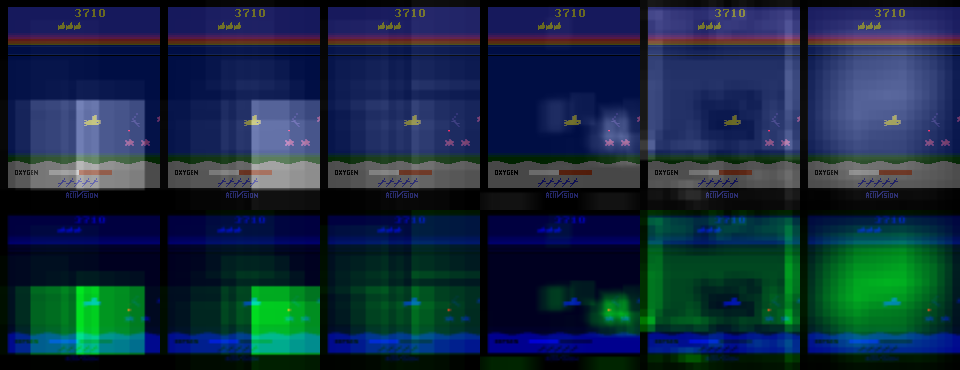}
        \label{fig:fixations:seaquest_2}
        \\ (d) & (e) & (f)
    \end{tabular}
    \caption{Atari-HEAD visualizations: (a,d) Breakout; (b,e) Enduro; (c,f) Seaquest. Each image shows the same frame with saliency maps produced by, left to right: DAQN~\cite{sorokin_darqn_2015}, RS-PPO~\cite{yang_ltiaa_2018}, \emph{Sparse FLS}, \emph{Dense FLS}, \emph{Sparse} + FLS after first conv layer, \emph{Sparse} + FLS after each conv layer. Human eye fixations are shown in red.}
    \label{fig:fixations}\vspace{-.3cm}
\end{figure*}

Fig.~\ref{fig:fixations} shows similar visualizations on the Atari-HEAD dataset. We have also made a full gameplay video available at~\url{https://youtu.be/i41rQXKsa50}.

\section{Conclusion}\label{sec:concl}

In this work, we have addressed the need for clear interpretable explanations for the behaviour of deep RL agents. We have proposed two new attention-based architectures designed to obtain saliency maps in the process of training while having competitive performance. Experiments on \emph{Atari} environments show that our architectures (both Sparse and Dense) allow to get interpretable visualizations and also exhibit several other advantages: the Sparse model with sum-pooling is smaller in terms of the number of parameters, while the Dense model provides the best visualizations.

We have studied two feature extractors for deep RL playing Atari games. The \emph{Sparse FLS} model achieves results very similar to the baseline but yields coarse visualizations. The \emph{Dense FLS} model, to the contrary, provides crisp images but achieves lower scores. One obvious open question is how to strike the balance between the two models we present. One plays well but yields worse visualizations, the other plays worse but produces excellent pictures~--- can we have the best of both worlds? Our results suggest that inserting an attention module between early convolutional layers where receptive fields are small or using multiple attention modules does not improve visualizations. One possible idea for further research would be to use multiple attention modules and maintain a loss term such as KL divergence to ensure that different modules yield similar attention maps. The attention modules could also be penalized with entropy loss. A visual attention module such as FLS could also improve interpretability in contexts other than deep reinforcement learning, such as image classification; this is also an avenue for further work. In general, we believe that visualizations can be further improved with custom loss functions, which again are a subject for further research.

A visual attention module such as FLS could also improve interpretability in contexts other than reinforcement learning, such as image classification; this is also an avenue for further work. We believe that continued work in this direction may improve state of the art in deep learning interpretability even further.

\FloatBarrier

{\small
\bibliographystyle{ieee}
\bibliography{egbib}
}

\section{Appendix}

\subsection{Performance details}

\begin{figure}[h]
    \centering
    \includegraphics[width=290pt,height=630pt,keepaspectratio]{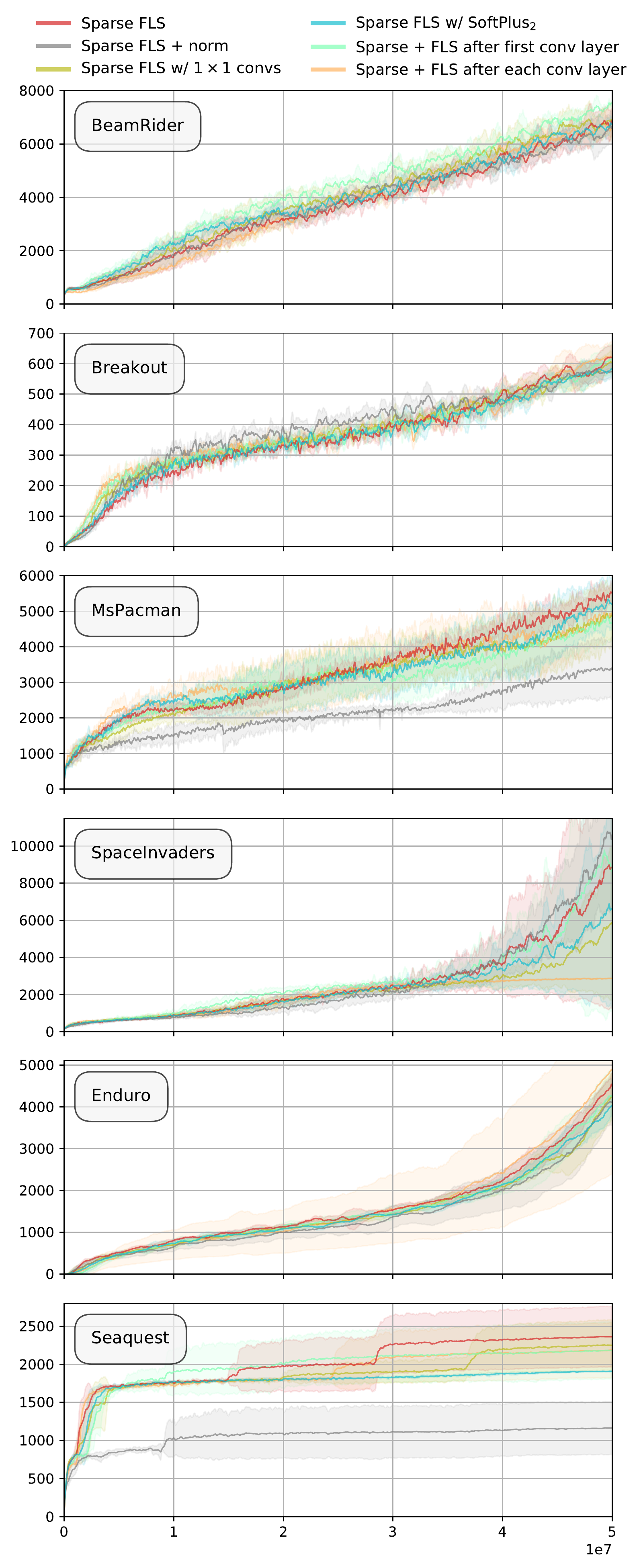}

    \caption{Reward curves for some models omitted in Fig.~\ref{fig:averaged_curves}, highlighting various modifications of Sparse FLS.}
    \label{fig:averaged_curves_other_1}
\end{figure}

\begin{figure}[h]
    \centering
    \includegraphics[width=290pt,height=630pt,keepaspectratio]{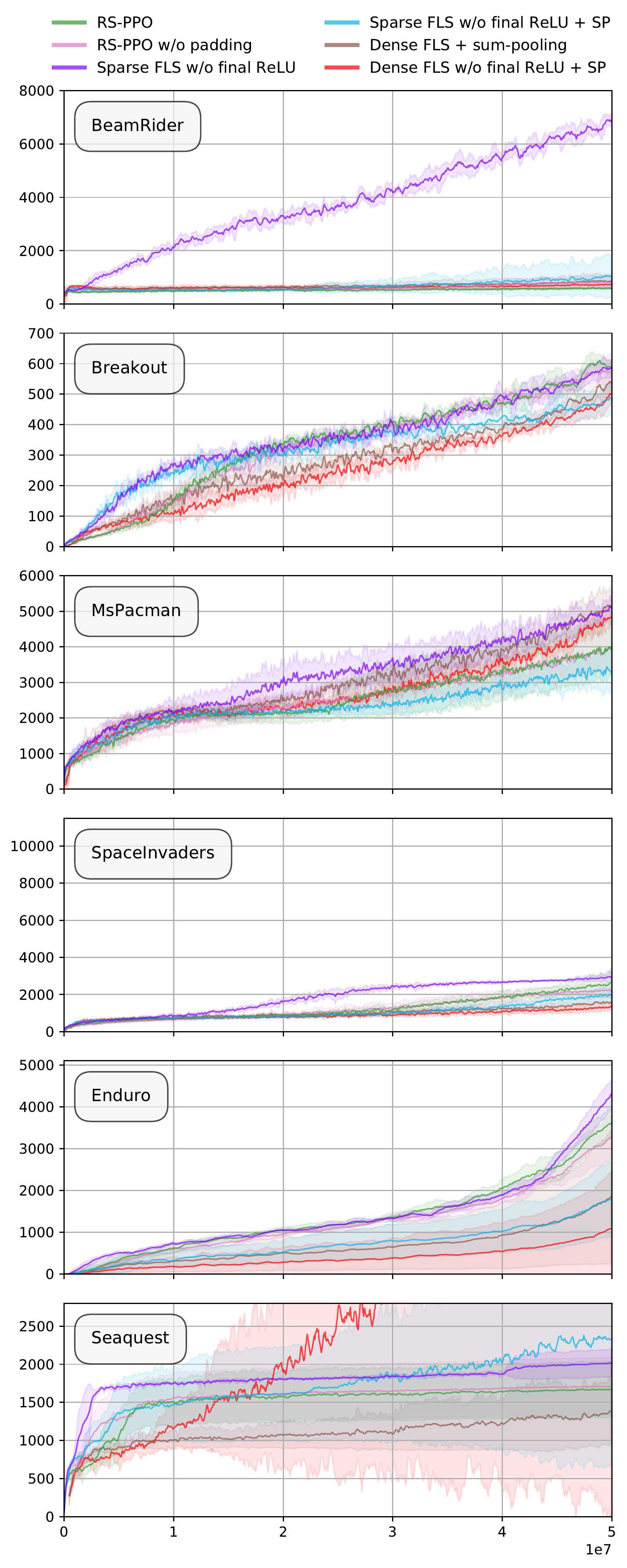}

    \caption{Reward curves for some models omitted in Fig.~\ref{fig:averaged_curves}. Note that RS-PPO w/o padding performs about as well as vanilla RS-PPO, and missing ReLU before attention destroys performance (except for Dense FLS on Seaquest).}
    \label{fig:averaged_curves_other_2}
\end{figure}

Figs.~\ref{fig:averaged_curves_other_1}--\ref{fig:averaged_curves_other_2} show curves for some models omitted in Fig.~\ref{fig:averaged_curves}.

\begin{figure}[h]
    \centering
    \includegraphics[width=300pt,height=630pt,keepaspectratio]{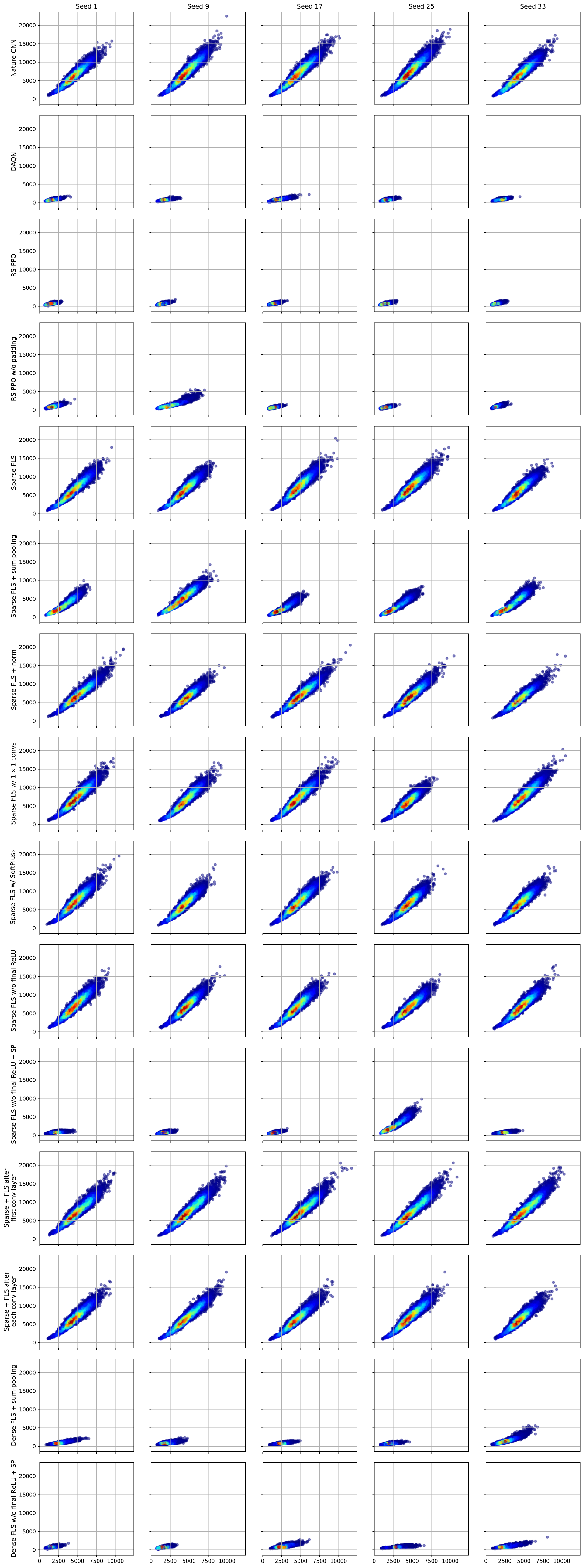}

    \caption{BeamRider scatterplot.}
    \label{fig:scatter:beamrider}
\end{figure}

\begin{figure}[h]
    \centering
    \includegraphics[width=300pt,height=630pt,keepaspectratio]{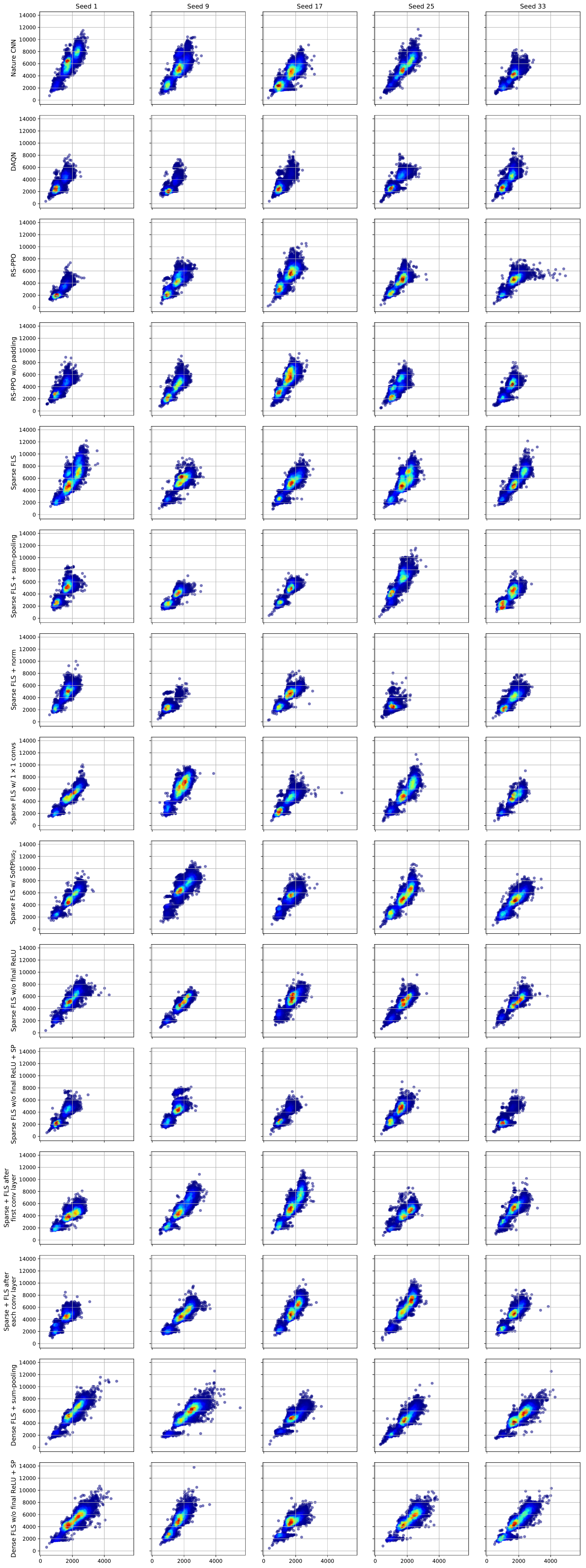}

    \caption{MsPacman scatterplot.}
    \label{fig:scatter:ms_pacman}
\end{figure}

\begin{figure}[h]
    \centering
    \includegraphics[width=300pt,height=630pt,keepaspectratio]{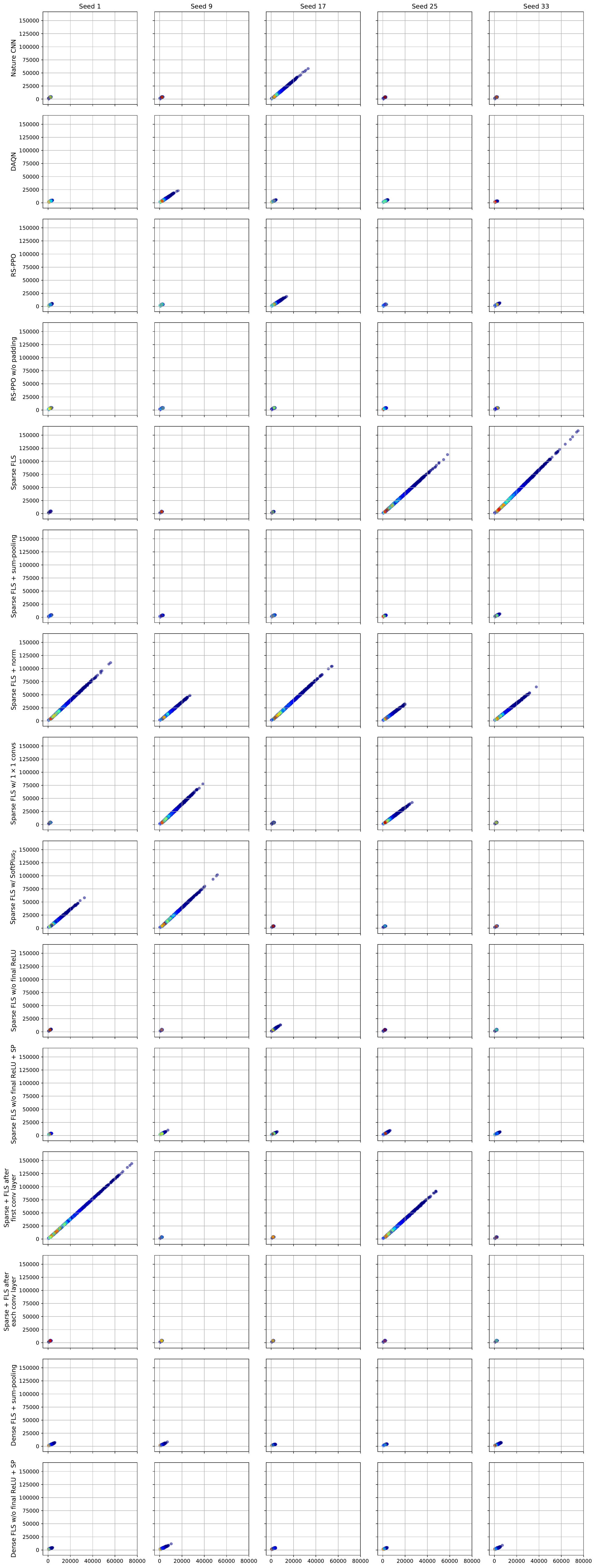}

    \caption{SpaceInvaders scatterplot.}
    \label{fig:scatter:space_invaders}
\end{figure}

\begin{figure}[h]
    \centering
    \includegraphics[width=300pt,height=630pt,keepaspectratio]{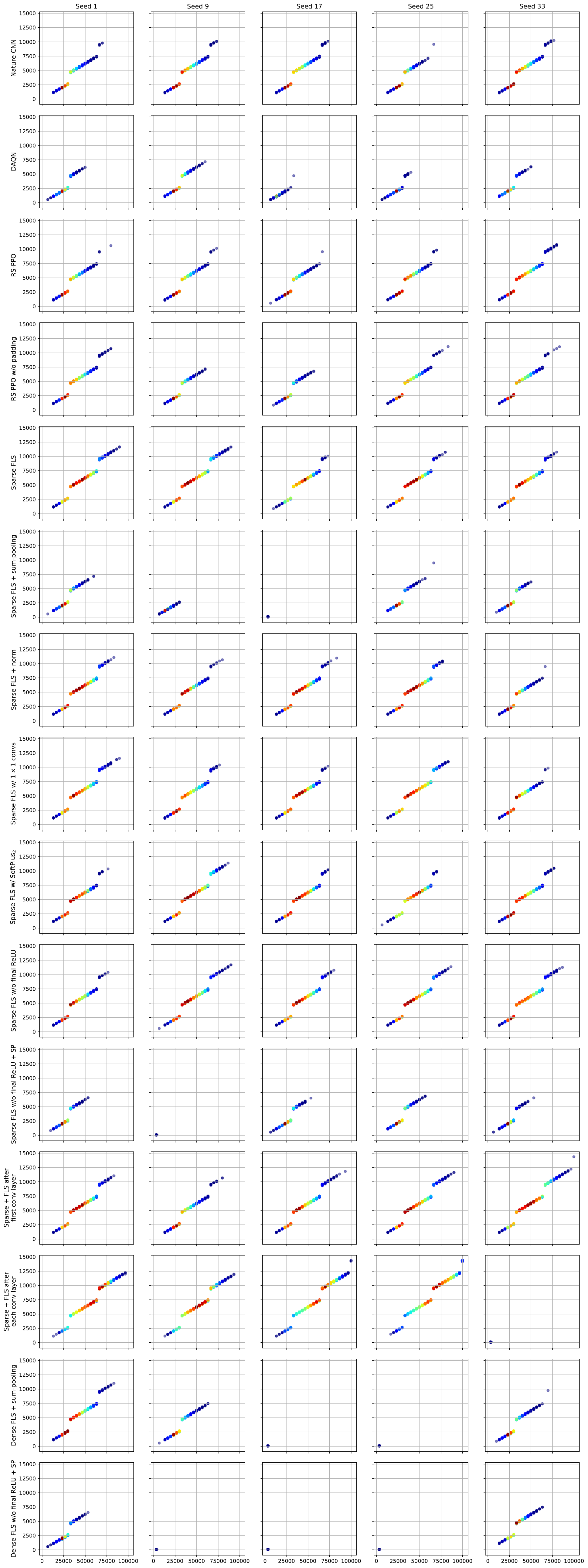}

    \caption{Enduro scatterplot.}
    \label{fig:scatter:enduro}
\end{figure}

\begin{figure}[h]
    \centering
    \includegraphics[width=300pt,height=630pt,keepaspectratio]{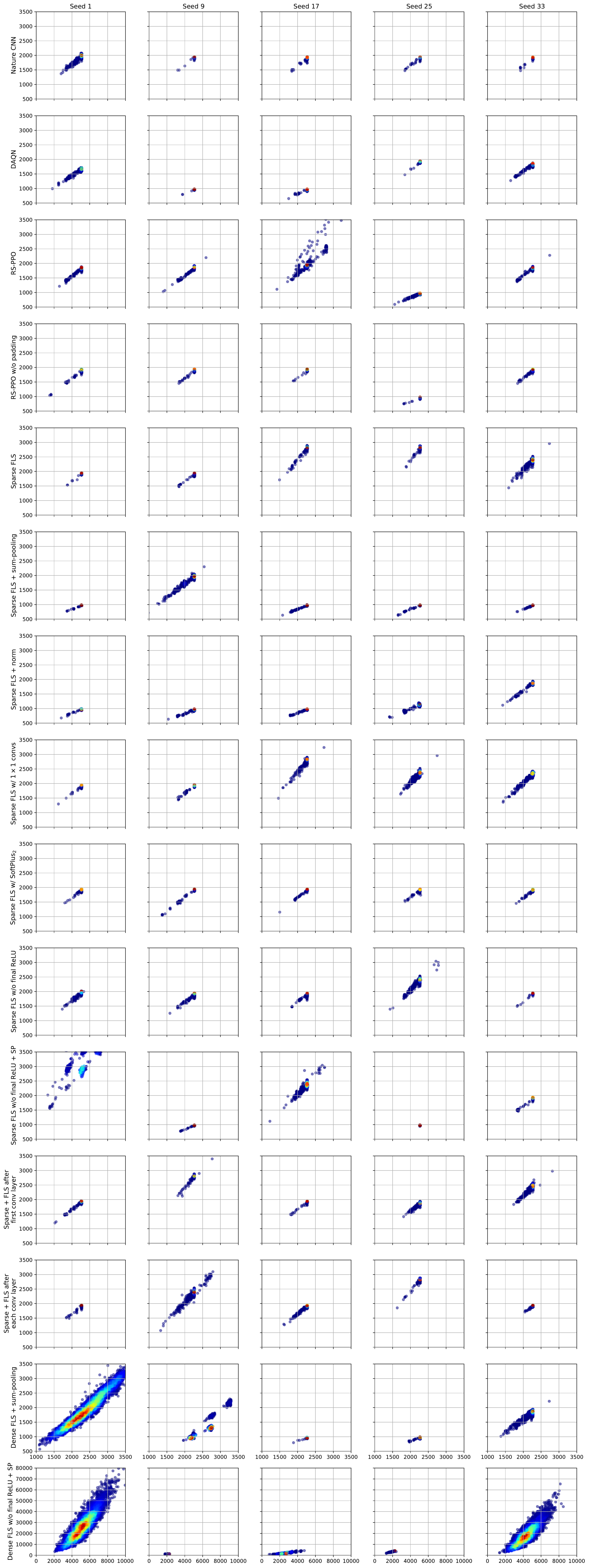}

    \caption{Seaquest scatterplot. Note the difference in scale on the vertical axis between the last model and the other ones.}
    \label{fig:scatter:seaquest}
\end{figure}

Figs.~\ref{fig:scatter:breakout}--\ref{fig:scatter:seaquest} show in detail performance evaluations summarized in Table~\ref{table:eval_scores}. Each scatterplot corresponds to one model. Each circle in each scatterplot corresponds to one completed episode. The horizontal axis shows the number of steps in the episodes; the vertical axis shows attained reward. Color is proportional to density.

\subsection{Breakout is censored}\label{sec:appendix_breakout}

\begin{figure}[h]
    \centering
    \includegraphics[width=300pt,height=630pt,keepaspectratio]{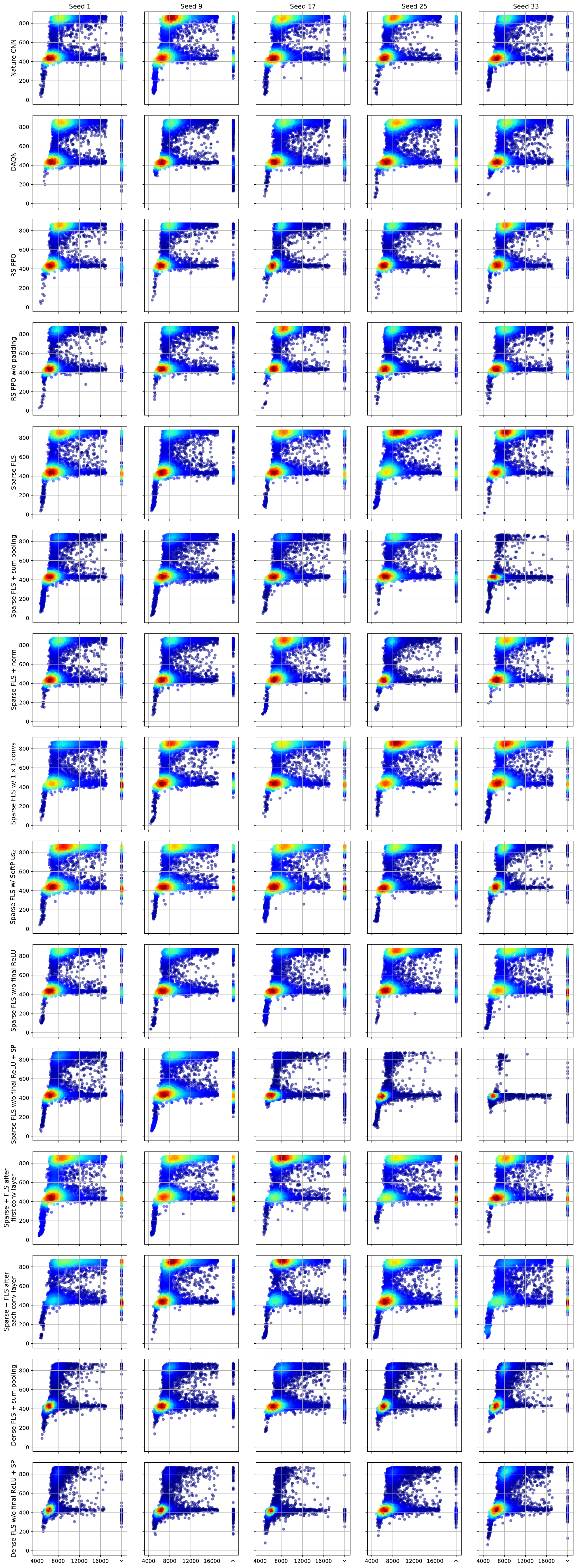}

    \caption{Breakout scatterplot.}
    \label{fig:scatter:breakout}
\end{figure}

\begin{figure}[h]
    \centering
    \includegraphics[width=300pt,height=630pt,keepaspectratio]{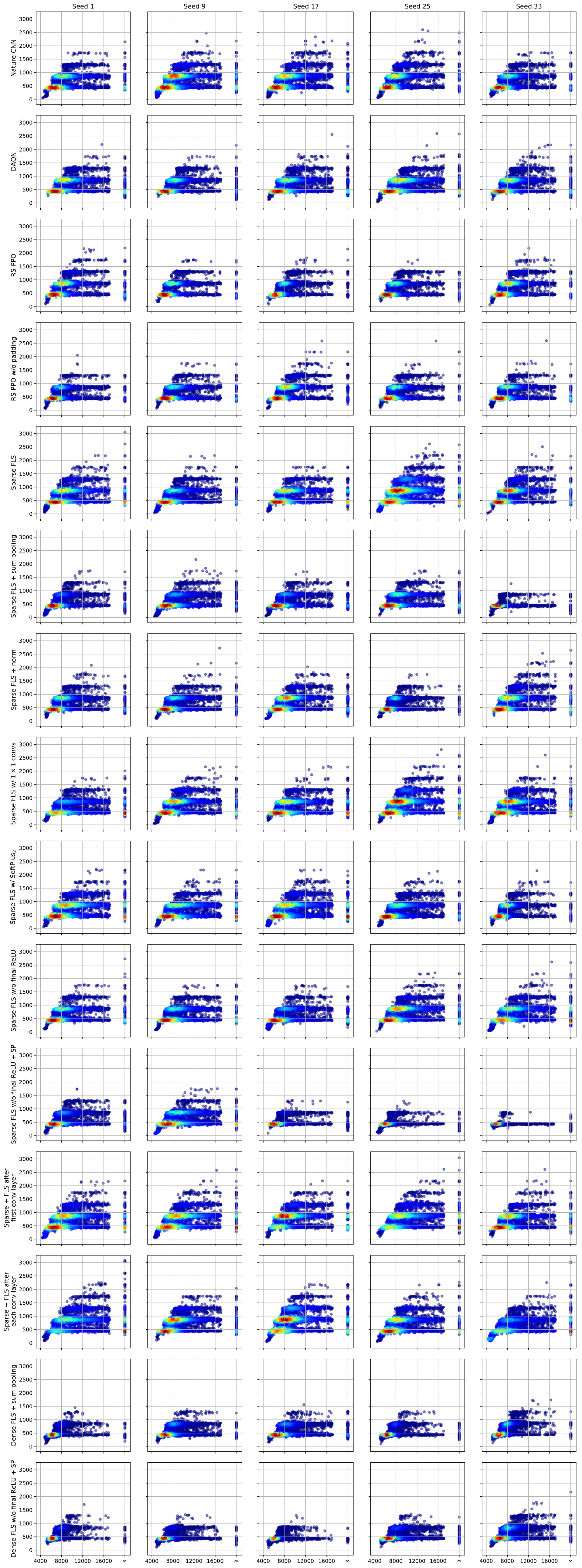}

    \caption{BreakoutInfinite scatterplot. See \ref{sec:appendix_breakout} for more details.}
    \label{fig:scatter:breakout_infinite}
\end{figure}

We pay special attention to Breakout (Fig.~\ref{fig:scatter:breakout}--\ref{fig:scatter:breakout_infinite}) because it is a popular benchmark. During experimentation, we noticed that none of the episodes attained a score greater than 864. It turned out that this was by design; the blocks could be cleared out only twice and did not respawn for the third time. This feature hampers our ability to compare agents that play the game; indeed, performance approaches 864 asymptotically, while variance remains high, making differences between models imperceptible. To work around this issue, we modified the original game so that blocks would respawn each time they are cleared and called the modified version ``BreakoutInfinite''. We did so by patching the \texttt{atari-py} wrapper from OpenAI to overwrite the score value in the emulator RAM: as soon as the value of 864 is attained, we replace it with 432, which triggers the built-in game logic for respawning blocks. We evaluated all of our models on the modified environment in addition to the standard one. The results are shown in Table~\ref{table:breakout_eval_scores}.

\begin{table}[h] \footnotesize\centering
    \begin{tabular}{p{3.9cm}p{1.0cm}p{2.1cm}}
        \toprule
        Game                                      & Breakout & BreakoutInfinite \\
        \midrule
        \rowcolor{tableau-c1}
        Nature CNN \cite{mnih_nature_2015}        &  618$\pm$209 &   652$\pm$274 (+5.4\%) \\
        \rowcolor{tableau-c2}
        DAQN \cite{sorokin_darqn_2015}            &  601$\pm$201 &   622$\pm$245 (+3.5\%) \\
        \rowcolor{tableau-c3}
        RS-PPO \cite{yang_ltiaa_2018}             &  605$\pm$202 &   625$\pm$244 (+3.3\%) \\
        RS-PPO \cite{yang_ltiaa_2018} w/o padding &  591$\pm$199 &   606$\pm$234 (+2.5\%) \\
        \midrule
        \rowcolor{tableau-c4}
        Sparse FLS                                &  624$\pm$211 &   663$\pm$283 (+6.4\%) \\
        \rowcolor{tableau-c5}
        Sparse FLS + sum-pooling                  &  520$\pm$183 &   529$\pm$204 (+1.6\%) \\
        Sparse FLS + norm                         &  598$\pm$200 &   621$\pm$247 (+3.8\%) \\
        Sparse FLS w/ $1 \times 1$ convs          &  621$\pm$207 &   650$\pm$272 (+4.8\%) \\
        Sparse FLS w/ $\operatorname{SoftPlus}_2$ &  612$\pm$208 &   641$\pm$268 (+4.8\%) \\
        Sparse FLS w/o final ReLU                 &  589$\pm$207 &   614$\pm$255 (+4.4\%) \\
        Sparse FLS w/o final ReLU + SP   &  480$\pm$158 &   484$\pm$172 (+0.9\%) \\
        Sparse + FLS after first conv layer       &  640$\pm$212 &   689$\pm$291 (+7.6\%) \\
        Sparse + FLS after each conv layer        &  633$\pm$217 &   681$\pm$302 (+7.7\%) \\
        \rowcolor{tableau-c6}
        Dense FLS + sum-pooling                   &  532$\pm$173 &   534$\pm$182 (+0.3\%) \\
        Dense FLS w/o final ReLU + SP    &  503$\pm$162 &   506$\pm$171 (+0.6\%) \\
        \bottomrule
    \end{tabular}\vspace{-.1cm}

    \caption{Comparison of evaluation scores for original and modified Breakout. SP denotes sum-pooling. Colors correspond to Fig.~\ref{fig:averaged_curves}.}\vspace{-0.4cm}
    \label{table:breakout_eval_scores}
\end{table}

We see that, as expected, in all cases the models perform better. However, the difference is not overwhelming, and never exceeds 8\%. Although we have not tested this, we hypothesize that training on ``BreakoutInfinite'' may improve model performance.

\makeatletter
\setlength{\@fptop}{0pt}
\makeatother

\end{document}

%% file: fig_architectures.tex
\begin{figure*}
	\centering
	\begin{minipage}{.54\linewidth}\centering
	\scalebox{\picscale}{
		\begin{tikzpicture}
		    \node [align=center]      (st) {$\cdots$};
			\node [conv, right=of st] (c1) {Conv \\ $32 \; @ \; 8 \times 8$ \\ s=4, p=0};
			\node [relu, right=of c1] (a1) {ReLU};
			\node [conv, right=of a1] (c2) {Conv \\ $64 \; @ \; 4 \times 4$ \\ s=2, p=0};
			\node [relu, right=of c2] (a2) {ReLU};
			\node [conv, right=of a2] (c3) {Conv \\ $64 \; @ \; 3 \times 3$ \\ s=1, p=0};
			\node [relu, right=of c3] (a3) {ReLU};
			\node [align=center, right=of a3] (vt) {$\cdots$};
			\node at (6.8,-1.2) {\Large (a)};

            \draw[->,thick] (st) -- (c1) node[midway, above] {$s_t$};
			\draw[->,thick] (c1) -- (a1);
			\draw[->,thick] (a1) -- (c2);
			\draw[->,thick] (c2) -- (a2);
			\draw[->,thick] (a2) -- (c3);
			\draw[->,thick] (c3) -- (a3);
			\draw[->,thick] (a3) -- (vt) node[midway, above] {$v_t$};
		\end{tikzpicture}
		}
	\end{minipage}~\begin{minipage}{.45\linewidth}\centering
	\scalebox{\picscale}{
		\begin{tikzpicture}
		    \node [op]                                 (add) {$\bm{+}$};
			\node [conv, left=of add,
			       yshift=4ex, minimum width=13ex]     (h1) {Conv \\ $N \; @ \; 1 \times 1$ \\ s=1, p=0};
		    \node [align=center, left=of h1]           (vt) {$\cdots$};
		    \node [block, fill=lightgray!25, left=of add,
		           yshift=-4ex, minimum width=13ex]    (linear) {Linear \\ w/o bias};
		    \node [align=center, left=of linear]       (ht) {$\cdots$};
			\node [actv, fill=purple!25, right=of add] (a1) {Tanh};
			\node [conv, right=of a1]                  (h2) {Conv \\ $1 \; @ \; 1 \times 1$ \\ s=1, p=0};
			\node [softmax, right=of h2]               (a2) {Spatial \\ softmax};
			\node [align=center, right=of a2]          (at) {$\cdots$};
						\node at (1.1,-1.1) {\Large (d)};

            \draw[->,thick] (vt) -- (h1) node[midway, above] {$v_t$};
            \draw[->,thick] (ht) -- (linear) node[midway, below] {$h_t$};
			\draw[->,thick] (h1) -- (add);
			\draw[->,thick] (linear) -- (add);
			\draw[->,thick] (add) -- (a1);
			\draw[->,thick] (a1) -- (h2);
			\draw[->,thick] (h2) -- (a2);
			\draw[->,thick] (a2) -- (at) node[midway, above] {$\alpha_t$};
		\end{tikzpicture}
		}
	\end{minipage}

	\begin{minipage}{.54\linewidth}\centering
		\scalebox{\picscale}{
		\begin{tikzpicture}
		    \node [align=center]      (st) {$\cdots$};
			\node [conv, right=of st] (c1) {Conv \\ $32 \; @ \; 7 \times 7$ \\ s=1, p=3};
			\node [relu, right=of c1] (a1) {ReLU};
			\node [conv, right=of a1] (c2) {Conv \\ $64 \; @ \; 5 \times 5$ \\ s=1, p=2};
			\node [relu, right=of c2] (a2) {ReLU};
			\node [conv, right=of a2] (c3) {Conv \\ $64 \; @ \; 3 \times 3$ \\ s=1, p=1};
			\node [relu, right=of c3] (a3) {ReLU};
			\node [align=center, right=of a3] (vt) {$\cdots$};
			\node at (6.8,-1.2) {\Large (b)};

            \draw[->,thick] (st) -- (c1) node[midway, above] {$s_t$};
			\draw[->,thick] (c1) -- (a1);
			\draw[->,thick] (a1) -- (c2);
			\draw[->,thick] (c2) -- (a2);
			\draw[->,thick] (a2) -- (c3);
			\draw[->,thick] (c3) -- (a3);
			\draw[->,thick] (a3) -- (vt) node[midway, above] {$v_t$};
		\end{tikzpicture}
		}
		\end{minipage}~\begin{minipage}{.45\linewidth}\centering
		\scalebox{\picscale}{
		\begin{tikzpicture}
		    \node [op]                                 (add) {$\bm{+}$};
			\node [conv, left=of add,
			       yshift=4ex, minimum width=13ex]     (h1) {Conv \\ $64 \; @ \; 1 \times 1$ \\ s=1, p=0};
		    \node [align=center, left=of h1]           (vt) {$\cdots$};
		    \node [block, fill=lightgray!25, left=of add,
		           yshift=-4ex, minimum width=13ex]    (linear) {Linear \\ w/o bias};
		    \node [align=center, left=of linear]       (ht) {$\cdots$};
			\node [actv, fill=purple!25, right=of add] (a1) {Tanh};
			\node [softmax, right=of a1]               (a2) {Spatial \\ softmax};
			\node [align=center, right=of a2]          (at) {$\cdots$};
			\node at (.5,-.8) {\Large (e)};

            \draw[->,thick] (vt) -- (h1) node[midway, above] {$v_t$};
            \draw[->,thick] (ht) -- (linear) node[midway, below] {$h_t$};
			\draw[->,thick] (h1) -- (add);
			\draw[->,thick] (linear) -- (add);
			\draw[->,thick] (add) -- (a1);
			\draw[->,thick] (a1) -- (a2);
			\draw[->,thick] (a2) -- (at) node[midway, above] {$\alpha_t$};
		\end{tikzpicture}
		}
		\end{minipage}

		\begin{minipage}{.54\linewidth}\centering
		\scalebox{\picscale}{
		\begin{tikzpicture}
		    \node [align=center]                      (vt) {$\cdots$};
			\node [conv, right=of vt]                 (h1) {Conv \\ $512 \; @ \; 1 \times 1$ \\ s=1, p=0};
			\node [actv, fill=yellow!25, right=of h1] (a1) {ELU};
			\node [conv, right=of a1]                 (h2) {Conv \\ $2 \; @ \; 1 \times 1$ \\ s=1, p=0};
			\node [softmax, right=of h2]  (a2) {Spatial \\ softmax};
			\node [pooling, right=of a2]              (pool) {Sum-pool \\ channels};
			\node [align=center, right=of pool]       (at) {$\cdots$};
			\node at (6.8,-1.2) {\Large (c)};

            \draw[->,thick] (vt) -- (h1) node[midway, above] {$v_t$};
			\draw[->,thick] (h1) -- (a1);
			\draw[->,thick] (a1) -- (h2);
			\draw[->,thick] (h2) -- (a2);
			\draw[->,thick] (a2) -- (pool);
			\draw[->,thick] (pool) -- (at) node[midway, above] {$\alpha_t$};
		\end{tikzpicture}
		}
		\end{minipage}~\begin{minipage}{.45\linewidth}\centering
		\scalebox{\picscale}{
		\begin{tikzpicture}
		    \node [align=center]                    (vt) {$\cdots$};
			\node [conv, right=of vt]               (h1) {Conv \\ $256 \; @ \; 3 \times 3$ \\ s=1, p=1};
			\node [relu, right=of h1]               (a1) {ReLU};
			\node [conv, right=of a1]               (h2) {Conv \\ $1 \; @ \; 3 \times 3$ \\ s=1, p=1};
			\node [actv, fill=lime!25, right=of h2] (a2) {Softplus};
			\node [align=center, right=of a2]       (at) {$\cdots$};
			\node at (5.8,-1.2) {\Large (f)};

            \draw[->,thick] (vt) -- (h1) node[midway, above] {$v_t$};
			\draw[->,thick] (h1) -- (a1);
			\draw[->,thick] (a1) -- (h2);
			\draw[->,thick] (h2) -- (a2);
			\draw[->,thick] (a2) -- (at) node[midway, above] {$\alpha_t$};
		\end{tikzpicture}
		}
		\end{minipage}

	\caption{Convolutional and attention blocks. At time $t$, input frames $s_t$ are turned into embeddings $v_t$, $h_t$ is the recurrent state: (a) Sparse convolutional block; (b) Dense convolutional block; (c) Region-Sensitive Module~\cite{yang_ltiaa_2018}; (d) DARQN~\cite{sorokin_darqn_2015} ($N = 256$) and~\cite{chen_observe_2017} ($N = 512$); (e) architecture from~\cite{mousavi_where_to_look_2016}; (f) our architecture ($h_t\equiv 0$).}
	\label{fig:conv}
\end{figure*}

\begin{figure}\centering
		\scalebox{\picscale}{
		\begin{tikzpicture}
			\node at (-1.5,0) {\Large (a)};
    		\node []                               (input) {};
			\node [composite, right=of input]      (conv) {Sparse block};
			\node [pooling, right=of conv]         (flatten) {Flatten};
			\node [align=center, right=of flatten] (etc) {$\cdots$};

            \draw[->,thick] (input) -- (conv) node[midway, above] {$s_t$};
			\draw[->,thick] (conv) -- (flatten);
			\draw[->,thick] (flatten) -- (etc);
		\end{tikzpicture}
		}\\[2pt]

		\scalebox{\picscale}{
		\begin{tikzpicture}
    		\node at (0.5,1) {\Large (b)};
    		\node []                          (input) {};
			\node [composite, right=of input] (conv) {Sparse block};
			\node [fork, right=of conv]       (fork) {};

			\node [composite, above right=of fork, yshift=0.05ex] (attn) {Attn};

			\node [op, below right=of attn, yshift=2.1ex]        (mul) {$\bm{\times}$};
			\node [pooling, right=of mul]          (flatten) {Spatial sum-pool};
			\node [align=center, right=of flatten] (etc) {$\cdots$};

            \draw[->,thick] (input) -- (conv) node[midway, above] {$s_t$};
			\draw[->,thick] (conv) -- (fork);
			\draw[->,thick] (fork) |- (attn);
			\draw[->,thick] (attn) -| (mul);
			\draw[->,thick] (fork) -- (mul);
			\draw[->,thick] (mul) -- (flatten);
			\draw[->,thick] (flatten) -- (etc);
		\end{tikzpicture}
		}\\[2pt]

		\scalebox{0.47}{
		\begin{tikzpicture}
			\node at (0.5,1) {\LARGE (c)};
    		\node []                                         (input) {};
    		\node [block, fill=lightgray!25, right=of input] (pad)  {Pad 1px};
			\node [composite, right=of pad]                  (conv) {Sparse block};
			\node [block, fill=lightgray!25, right=of conv]  (norm) {$L^2$ norm};
			\node [fork, right=of norm]                      (fork) {};

			\node [composite, above right=of fork, yshift=2.25ex] (attn) {RS module};

			\node [op, below right=of attn]        (mul) {$\bm{\times}$};
			\node [pooling, right=of mul]          (flatten) {Flatten};
			\node [align=center, right=of flatten] (etc) {$\cdots$};

            \draw[->,thick] (input) -- (pad) node[midway, above] {$s_t$};
            \draw[->,thick] (pad) -- (conv);
			\draw[->,thick] (conv) -- (norm);
			\draw[->,thick] (norm) -- (fork);
			\draw[->,thick] (fork) |- (attn);
			\draw[->,thick] (attn) -| (mul);
			\draw[->,thick] (fork) -- (mul);
			\draw[->,thick] (mul) -- (flatten);
			\draw[->,thick] (flatten) -- (etc);
		\end{tikzpicture}
		}\\[2pt]

		\scalebox{0.53}{
		\begin{tikzpicture}
			\node at (0.5,1) {\LARGE (d)};
    		\node []                          (input) {};
			\node [composite, right=of input] (conv) {Sparse or dense block};
			\node [fork, right=of conv]       (fork) {};

			\node [composite, above right=of fork, yshift=2.25ex] (attn) {FLS module};

			\node [op, below right=of attn]        (mul) {$\bm{\times}$};
			\node [pooling, right=of mul]          (flatten) {Flatten or \\ spatial sum-pool};
			\node [align=center, right=of flatten] (etc) {$\cdots$};

            \draw[->,thick] (input) -- (conv) node[midway, above] {$s_t$};
			\draw[->,thick] (conv) -- (fork);
			\draw[->,thick] (fork) |- (attn);
			\draw[->,thick] (attn) -| (mul);
			\draw[->,thick] (fork) -- (mul);
			\draw[->,thick] (mul) -- (flatten);
			\draw[->,thick] (flatten) -- (etc);
		\end{tikzpicture}
		}

	\caption{Model architectures; $s_t$ -- input frames at time $t$: (a) Nature CNN \cite{mnih_nature_2015}; (b) DAQN, inspired by \cite{sorokin_darqn_2015}; (c) LTIAA \cite{yang_ltiaa_2018}; (d) our architecture. Fig.~\ref{fig:conv} shows the Sparse and Dense blocks and attention modules.}\vspace{-.2cm}
	\label{fig:architectures}
\end{figure}